%% file: main.tex
\documentclass[10pt,twocolumn,letterpaper]{article}
\usepackage{cvpr}
\usepackage[dvipsnames]{xcolor}
\usepackage{comment}
\usepackage{bm}
\usepackage{algorithm}
\usepackage{algpseudocode}
\usepackage{multirow}
\usepackage{pifont}

\definecolor{cvprblue}{rgb}{0.21,0.49,0.74}
\usepackage[pagebackref,breaklinks,colorlinks,citecolor=cvprblue]{hyperref}

\newcommand{\method}{\textsc{Shap-Editor}\xspace} 

\newcommand{\x}{\bm{x}}

\newcommand{\br}{\bm{r}}

\newcommand{\bd}{\bm{d}}
\newcommand{\beps}{\bm{\epsilon}}
\DeclareMathOperator*{\E}{\mathbb{E}}
\newcommand{\xmark}{\ding{55}}%
\newcommand{\cmark}{\ding{51}}%

\newcommand{\titi}{TI2I}  %
\newcommand{\iptp}{\hat{\beps}_\text{\titi}}
\newcommand{\tti}{\hat{\beps}_\text{T2I}}
\makeatletter
\renewcommand{\paragraph}{%
  \@startsection{paragraph}{4}%
  {\z@}{0.5em}{-1em}%
  {\normalfont\normalsize\bfseries}%
}
\makeatother

\def\paperID{1870} %
\def\confName{CVPR}
\def\confYear{2024}

\title{\method: Instruction-guided Latent 3D Editing in Seconds}

\author{
Minghao Chen
\quad
Junyu Xie
\quad
Iro Laina
\quad
Andrea Vedaldi
\\[0.3em]
Visual Geometry Group, University of Oxford\\
{\tt\small \{minghao, jyx, iro, vedaldi\}@robots.ox.ac.uk} \\
\href{https://silent-chen.github.io/Shap-Editor/}{\tt\small {\nolinkurl{silent-chen.github.io/Shap-Editor}}}
}
\begin{document}

\twocolumn[{
\renewcommand\twocolumn[1][]{#1}%
\maketitle
\begin{center}
\centering
\includegraphics[width=\linewidth]{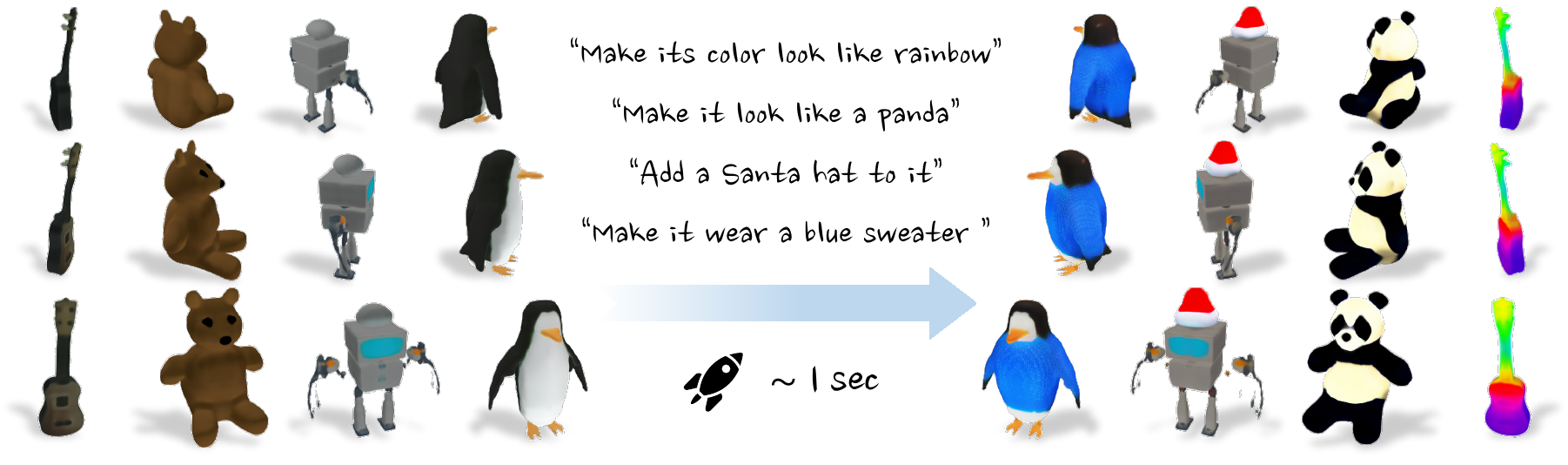}
\vspace{-0.35cm}
\captionsetup{type=figure}
\captionof{figure}{Given 3D assets as inputs, \method achieves fast editing within one second by learning a feed-forward mapping in the latent space of a 3D asset generator.
}\label{fig:teaser}
\vspace{0.64cm}
\end{center}
}]
\input{sec/0_abstract}
\input{sec/1_intro}
\input{sec/2_related}

\input{sec/3_method}

\input{sec/4_results}

\input{sec/5_conclusion}

\paragraph{Ethics.}

We use the OmniObject3D dataset following their terms and conditions.
This data contains no personal data.
For further details on ethics, data protection, and copyright, please see \url{https://www.robots.ox.ac.uk/~vedaldi/research/union/ethics.html}.

\paragraph{Acknowledgements.}
This research is supported by ERC-CoG UNION 101001212.
I.~L.~is also partially supported by the VisualAI EPSRC grant (EP/T028572/1).
J.~X.~ is supported by the Clarendon Scholarship.
We also appreciate the valuable discussions and support from Paul Engstler, Tengda Han, Laurynas Karazija, Ruining Li, Luke Melas-Kyriazi, Christian Rupprecht, Stanislaw Szymanowicz, Jianyuan Wang, Chuhan Zhang, Chuanxia Zheng, and Andrew Zisserman. 

{
\small
\bibliographystyle{ieeenat_fullname}
\bibliography{references}
}
\clearpage
\input{sec/supp}

\end{document}

%% file: sec/0_abstract.tex
\begin{abstract}
\vspace{-0.1cm}
We propose a novel feed-forward 3D editing framework called \emph{\method}.
Prior research on editing 3D objects primarily concentrated on editing individual objects by leveraging off-the-shelf 2D image editing networks.
This is achieved via a process called distillation, which transfers knowledge from the 2D network to 3D assets.
Distillation necessitates at least tens of minutes per asset to attain satisfactory editing results, and is thus not very practical.
In contrast, we ask whether 3D editing can be carried out directly by a feed-forward network, eschewing test-time optimization.
In particular, we hypothesise that editing can be greatly simplified by first encoding 3D objects in a suitable latent space.
We validate this hypothesis by building upon the latent space of Shap-E.
We demonstrate that direct 3D editing in this space is possible and efficient by building a feed-forward editor network that only requires approximately one second per edit.
Our experiments show that \emph{\method} generalises well to both in-distribution and out-of-distribution 3D assets with different prompts, exhibiting comparable performance with methods that carry out test-time optimisation for each edited instance.
\end{abstract}

%% file: sec/1_intro.tex
\vspace{-0.5cm}
\section{Introduction}%
\label{sec:intro}

We consider the problem of generating and editing 3D objects based on instructions expressed in natural language.
With the advent of denoising diffusion models~\cite{sohl2015deep,ho2020denoising,song2021denoising,rombach2022high}, text-based image generation~\cite{rombach2022high} and editing~\cite{brooks2023ip2p,hertz2023prompttoprompt,mokady2023null} have witnessed remarkable progress.
Many authors have since attempted to transfer such capabilities to 3D via \emph{test-time optimisation}, where a 3D model is optimised from scratch until its rendered 2D appearance satisfies an underlying prior in the pre-trained 2D models~\cite{instructnerf2023,kamata2023instruct,Sella_2023_ICCV}.

While optimisation-based methods obtain encouraging results, they are not scalable\,---\,in fact, a single 3D generation or edit can take from minutes to hours.
It is then natural to seek for more efficient generators and editors that can directly work in 3D.
We hypothesise that this can be greatly facilitated by first learning a suitable \emph{latent space} for 3D models.
For instance, Shape-E~\cite{jun2023shap} has recently learned an auto-encoder that maps 3D objects into vectors (latents). These vectors can be generated \emph{directly} by a diffusion model, eschewing test-time optimisation entirely.

In this paper, we thus ask whether such a 3D latent space can support not only efficient 3D generation, but also efficient 3D editing.
We answer affirmatively by developing a method, \method, that can apply semantic, text-driven edits directly in the latent space of 3D asset generators.
Because of the properties of the latent space, once learned, the editor function is capable of applying the edit to any new object in just one second \emph{vs.}~minutes to hours required by optimisation-based approaches.

In more detail, our method starts from a \emph{3D auto-encoder}\,---\,\eg, the off-the-shelf Shape-E encoder.
It also takes as input a \emph{2D image editor} that can understand instructions in natural language.
For any such instruction, \method learns a  function that can map, in a feed-forward manner, the latent of any 3D object into the latent of the corresponding edit\,---\,we call this a \emph{latent editor}.
The fact that the latent editor can be learned relatively easily is a strong indication that the 3D latent space has a useful structure for this type of operations.
Empirically, we further explore and demonstrate the partial linearity of such edits when they are carried out in this space.

Our method has several interesting practical properties.
First, we learn a single latent editor that works universally for any input object.
This function lifts to 3D space the knowledge contained in the 2D image editor via distillation losses.
In fact, we show that we can distill simultaneously \emph{several} different 2D editors of complementary strengths.
In our student-teacher framework, the \emph{combined} knowledge of the editors is then transferred to the latent editor.

Second, we note that the latent editor is able to capture certain semantic concepts, and in particular complex compositions of concepts, better than the original text-to-3D generator.
Moreover, it allows the application of several edits sequentially, with cumulative effects.

Third, while our method learns an editor function for each type of edit, at test time it can be applied to any number of objects very quickly, which could be used to modify libraries of thousands of 3D assets (\eg, to apply a style to them).
In this sense, it can be seen as an amortised counterpart to methods that use test-time optimisation.
We also demonstrate that by conditioning the latent editor on text, several different edits can be learned successfully by a single model.
This suggests that, given sufficient training resources, it might be possible to learn an open-ended editor.

To summarise, our contributions are:
(1) We show that 3D latent representations of objects designed for generation can also support semantic editing;
(2) We propose a method that can distill the knowledge of one or more 2D image generators/editors in a single latent editor function which can apply an edit in seconds, significantly reducing the computational costs associated with test-time optimisation; %
(3) We show that this latent function does  better at compositional tasks than the original 3D generator;
(4) We further show that it is possible to extend the latent editor to understand multiple editing instructions simultaneously. %

%% file: sec/2_related.tex
\section{Related work}%
\label{sec:rw}

\paragraph{Diffusion-based image manipulation.}

Recent advances in text-guided diffusion models have greatly improved 2D image generation.
Yet, these models typically offer limited control over the generated content.
To enable controllable generation, researchers have explored concept personalisation~\cite{gal2022textual, ruiz2023dreambooth, kumari2022customdiffusion}, layout control~\cite{li2023gligen, chefer2023attendandexcite, chen2023trainingfree, epstein2023selfguidance}, and other conditionings~\cite{Zhang_2023_ICCV}. Other recent works~\cite{meng2022sdedit, mokady2023null, bar2022text2live, Tumanyan_2023_CVPR, parmar2023zeroshot, kawar2023imagic, hertz2023prompttoprompt} have extended text-guided diffusion models to image-to-image translation tasks and image editing.
InstructPix2Pix (IP2P)~\cite{brooks2023ip2p} finetunes a diffusion model to accept image conditions and instructional prompts as inputs, by training on a large-scale synthetic dataset. 
Subsequent research~\cite{Zhang2023MagicBrush, zhang2023hive} has sought to further finetune InstructPix2Pix with manually annotated datasets.

\paragraph{Neural field manipulation.}

Several attempts have been made to extend neural fields, such as NeRFs~\cite{mildenhall2021nerf}, with editing capabilities.
EditNeRF~\cite{liu2021editing} was the first approach to edit the shape and color of a NeRF given user scribbles.
Approaches that followed include 3D editing from just a single edited view~\cite{bao2023sine}, or via 2D sketches~\cite{mikaeili2023sked}, keypoints~\cite{zheng2023editablenerf}, attributes~\cite{kania2022conerf}, meshes~\cite{yuan2022nerf,xu2022deforming,peng2022cagenerf,yang2022neumesh,jambon2023nerfshop} or point clouds~\cite{chen2023neuraleditor}.
Others focus on object removal with user-provided points or masks~\cite{tschernezki2022neural,weder2023removing,mirzaei2023spin}, object repositioning~\cite{yang2021learning}, recoloring~\cite{kuang2023palettenerf, gong2023recolornerf,lee2023ice} and style transfer~\cite{zhang2022arf,xu2023desrf}.

\paragraph{Text-to-3D generation.}

Given the success of diffusion-based generation and vision-language models such as CLIP~\cite{radford2021learning}, several methods have been proposed for generating 3D scenes using text prompts~\cite{clipmesh, dreamfiled}.
A pioneering work is DreamFusion~\cite{poole2022dreamfusion}, which proposes the Score Distillation Sampling (SDS) loss.
They use it to optimise a parametric model, such as NeRF, with the supervision of an off-the-shelf 2D diffusion model.
DreamFusion has since been improved by followups~\cite{lin2023magic3d, metzer2023latent, wang2023prolificdreamer, chen2023fantasia3d}, but these methods are generally not directly applicable to 3D editing tasks. %
Another direction is to train auto-encoders on explicit 3D representation, such as point cloud \cite{zeng2022lion} or voxel grid \cite{sanghi2023clip}, or on implicit functions, such as singed distance function \cite{fu2022shapecrafter} or neural radiance field \cite{kosiorek2021nerf}. The generative models are trained on the latent space \cite{nichol2022point, chen2022transformers} conditioned on text inputs. The most related work is Shap-E \cite{jun2023shap} that trained on a very large-scale dataset (several million). It encodes 3D assets into latents and can directly output implicit functions, such as NeRFs, signed distance functions and texture fields~\cite{shen2021deep, gao2022get3d}. It also incorporates a diffusion model \cite{ho2020denoising} for the conditional 3D asset generation part.

\paragraph{Text-based 3D editing.}

Differently from text-to-3D generation, editing methods start from a given 3D object or scene (usually represented by a NeRF~\cite{mildenhall2021nerf} or voxel grid~\cite{sun2022direct}).
Some authors leverage CLIP embeddings or similar models~\cite{luddecke2022image,li2022languagedriven,li2022grounded} to perform text-driven semantic editing/stylisation globally~\cite{wang2022clip,wang2023nerf,michel2022text2mesh,lei2022tango} or locally~\cite{song2023blending,kobayashi2022decomposing, wang2023inpaintnerf360, gordon2023blended}.

Most recent and concurrent approaches leverage diffusion priors.
Starting with InstructNeRF2NeRF~\cite{instructnerf2023}, one line of research employs pre-trained 2D models to edit image renderings of the original model and uses these to gradually update the underlying 3D representation~\cite{instructnerf2023,yu2023edit,wang2023proteusnerf}.
Instead of editing images, others optimise the 3D representation directly with different variants of score distillation sampling~\cite{kamata2023instruct,Sella_2023_ICCV,zhuang2023dreameditor,park2023ed,li2023focaldreamer,cheng2023progressive3d,zhang2023text, zhou2023repaint}.
They often differ in their use of the diffusion prior; \eg,~\cite{instructnerf2023, kamata2023instruct} use InstructPix2Pix~\cite{brooks2023ip2p}, while most others rely on Stable Diffusion~\cite{ho2020denoising}.
Many existing methods edit scenes globally, which may sometimes affect unintended regions.
To address this issue, approaches such as Vox-E~\cite{Sella_2023_ICCV} and FocalDreamer~\cite{li2023focaldreamer}, introduce mechanisms for local 3D editing.
We note, however, that, due to their inherent design, most methods cannot handle global and local edits equally well.

In contrast, we show that we can train a \emph{single} network for \emph{both} types of edits with a loss tailored to each edit type.
We also note that all these methods perform editing via test-time optimisation, which does not allow interactive editing in practice;~\cite{yu2023edit,wang2023proteusnerf} focus on accelerating this process, but they still use an optimisation-based approach.
Instead, our feed-forward network applies edits instantaneously.

%% file: sec/3_method.tex
\section{Method}%
\label{sec:method}

\begin{figure*}
\centering
\includegraphics[width=0.95\linewidth]{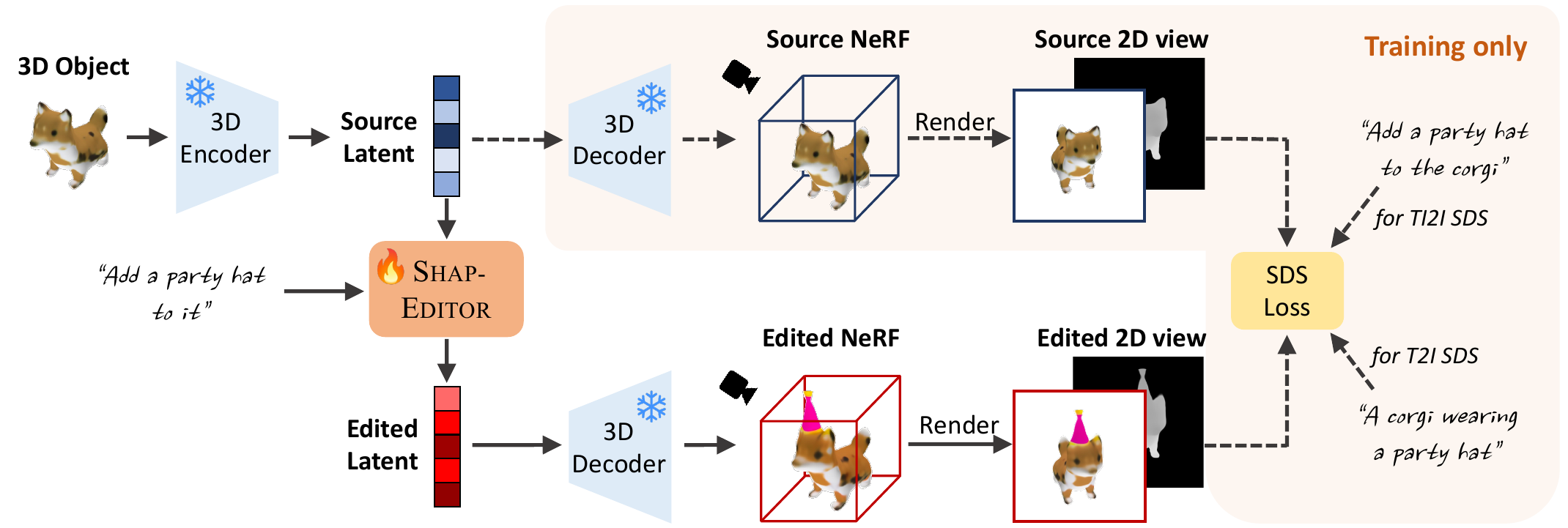}
\vspace{-0.05cm}
\caption{\textbf{Latent 3D editing with \method.} During training, we use the Shap-E encoder to map a 3D object into the latent space. The source latent and a natural language instruction are then fed into an editing network that produces an edited latent. The edited latent and original latent are decoded into NeRFs and we render a pair of views (RGB images and depth maps) with the same viewpoint for the two different NeRF. The paired views are used for distilling knowledge from the pre-trained 2D editors with our design training objective to our \method. During inference, one only needs to pass the latent code to our \method, resulting in fast editing.}%
\label{fig:pipeline}
\vspace{-0.2cm}
\end{figure*}

Let $\theta$ be a model of a 3D object, specifying its shape and appearance.
Common choices for $\theta$ include textured meshes and radiance fields, but these are often difficult to use directly in semantic tasks such as text-driven 3D generation and editing.
For images, generation and editing are often simplified by adopting a latent representation.
In this paper, we thus ask whether replacing $\theta$ with a corresponding latent code $\br$ can result in similar benefits for 3D editing.

More formally, we consider the problem of constructing an \emph{editor} function $f : (\theta^s,y) \mapsto \theta^e$ which takes as input a 3D object $\theta^s$ (source) and produces as output a new version of it $\theta^e$ (edit) according to natural-language instructions $y$.
For example, $\theta^s$ could be the 3D model of a \textit{corgi}, $y$ could say \textit{``Give it a Christmas hat''}, then $\theta^e$ would be the \textit{same corgi} but with the \textit{hat}.
Learning the map $f$ directly is challenging because interpreting natural language in an open-ended manner requires large models trained on billions of data samples, which are generally not available in 3D.

Some authors have approached this problem by starting from existing 2D image models, trained on billions of images.
We can think of a 2D editor as a conditional distribution $p(\x^e \! \mid \! \x^s,y)$ of possible edits $\x^e$ given the source image $\x^s$.
Then, one can obtain $\theta^e$ by optimising the log-posterior
$
\E_\pi\left[\log p(\mathcal{R}(\theta^e, \pi) \mid \x^s,y)\right]
$
where $\mathcal{R}(\theta^e, \pi)$  %
is the image obtained by rendering $\theta^e$ from a random viewpoint $\pi$ with a differentiable renderer $\mathcal{R}$.
This, however, requires \emph{per-instance optimisation at test time}, so obtaining $\theta^e$ may take minutes to hours in practice.

Here, we thus study the problem of learning a much faster \emph{feed-forward editor} function $f$. %
To do so, we first consider a pair of encoder-decoder functions $h: \theta \mapsto \br$ and $h^*: \br \mapsto \theta$, mapping the 3D object $\theta$ to a corresponding latent representation $\br$.
We then reduce the problem to learning a \emph{latent editor} $g : (\br^s,y) \mapsto \br^e$ which performs the \emph{edit directly in latent space}.
Hence, we decompose the editor as
$
f|_y = h^* \circ g|_y \circ h.
$
This can be advantageous if, by exploiting the compactness and structure of the latent space, the latent editor $g|_y$ can be fast, efficient, and easy to learn.

In the rest of the section, we review important background (\Cref{s:background}), explain how we build and train the latent editor 
(\Cref{s:edits}), and finally describe a combination of 2D priors for global and local edits (\Cref{s:2d-editors}). %

\subsection{Background}%
\label{s:background}

\paragraph{Shap-E\@: an off-the-shelf 3D latent space.}
Instead of learning a latent space from scratch, we turn to a pre-trained off-the-shelf model, Shap-E~\cite{jun2023shap}, which is a conditional generative model of 3D assets that utilises a latent space.
It comprises an auto-encoder that maps 3D objects to latent codes as well as a diffusion-based text/image-conditioned generator that operates in said space.
In our work, we mainly use the encoder/decoder components, denoted as $h$ and $h^*$, respectively, mapping the 3D object from/to a latent vector $\br \in \mathbb{R}^{1024 \times 1024}$.
In an application, the source latent $\br^s$ can be either obtained using $h$ starting from a mesh, or can be sampled from a textual description using the Shape-E generator.
For more details, please refer to the \Cref{sec:detailed_implementation}.

\paragraph{Score Distillation Sampling (SDS).}

SDS~\cite{poole2022dreamfusion} is a loss useful for distilling diffusion probabilistic models (DPMs).
Recall that a DPM models a data distribution $p(\x)$ by learning a denoising function
$
\beps \approx \hat{\beps} (\x_t; y, t),
$
where $\x_t = \alpha_t \x + \sigma_t \beps$ is a noised version of the data sample $\x$.
Here $(\alpha_t,\sigma_t)$ define the noise schedule, $\beps \sim \mathcal{N}(0,I)$ is normally distributed, and $t=0,1,\dots,T$ are noising steps.
The SDS energy function is given by
$
\mathcal{L}_\text{SDS}(\x)
=
\mathbb{E}_{t, \beps}
\Big[
- \sigma_t \log p(\x_t)
\Big]
$,
where $p(\x_t)$ is the noised version of the data distribution $p(\x)$ and the noise level is picked randomly according to a distribution $w(t)$.
The reason for choosing this distribution is that the denoising function is also an estimator of the gradient of the log-posterior $\log p(\x_t;y,t)$, in the sense that
$
\hat{\beps} (\x_t; y, t) = - \sigma_t \log p(\x_t; y,t) .
$
Hence, one obtains the gradient estimator
\begin{equation}\label{eq:sds_loss}
\nabla_{\x}
\mathcal{L}_{\text{SDS}}(\x)
=
\mathbb{E}_{t, \beps}
\Big[
 \hat{\beps} \,(\x_t; y, t)  - \beps
\Big]
\end{equation}
For 3D distillation, $\x = \mathcal{R}(\theta,\pi)$, so the chain rule is used to compute the gradient w.r.t.~$\theta$ and the loss is also averaged w.r.t.~random viewpoints $\pi$.

\subsection{3D editing in latent space}%
\label{s:edits}

We now consider the problem of learning the latent editor $g$ {(\ie, our \method)},  using the method summarised in \Cref{fig:pipeline,alg:3DE}.
Learning such a function would require suitable triplets $(\theta^s,\theta^e,y)$ consisting of source and target 3D objects and the instructions $y$, but there is no such dataset available.
Like prior works that use test-time optimisation, we start instead from an existing 2D editor, implementing the posterior distribution $p(\x^e \!\mid \! \x^s,y)$,
but we only use it for supervising $g$ at training time, not at test time.
A benefit is that this approach can fuse the knowledge contained in different 2D priors into a single model, which, as we show later, may be better suited for different kinds of edits (\eg, local vs global).

\paragraph{Training the latent editor.}%
\label{s:training}

Training starts from a dataset $\Theta$ of source 3D objects $\theta^s$ which are then converted in corresponding latent codes $\br^s = h(\theta^s)$ by utilising the encoder function $h$ or sampling the text-to-3D generator $p(\br^s \mid y^s)$ given source descriptions $y^s$.

The latent editor $\br^e = g(\br^s,y)$ is tasked with mapping the source latent $\br^s$ to an edited latent $\br^e$ based on instructions $y$.
We supervise this function with a 2D editor (or mixture of editors) providing the conditional distribution $p(\x^e \! \mid \! \x^s,y)$.
Specifically, we define a loss of the form:
\begin{equation}
\mathcal{L}_\text{SDS-E}(\x^e \! \mid \! \x^s,y)
=
\mathbb{E}_{t,\beps}
\left[
- \sigma_t \log p(\x^e_t \mid \x^s,y)
\right],
\end{equation}
where
$\x^e_t = \alpha_t \x^e + \sigma_t \beps$, and
$\x^s = \mathcal{R}(h^*(\br^s),\pi)$ and
$\x^e = \mathcal{R}(h^*(\br^e),\pi)$ are renders of the object latents $\br^s$ and $\br^e$, respectively, from a randomly-sampled viewpoint $\pi$.
Importantly, the rendering functions are differentiable.

We choose this loss because its gradient can be computed directly from any DPM implementation of the 2D editor (\Cref{s:background}).
At every learning iteration, a new source latent $\br^s$ is considered, the edited image $\x^e = \mathcal{R}(g(\br^s,y), \pi)$ is obtained, and the gradient
$
\nabla_{\x_e}\mathcal{L}_\text{SDS-E}(\x^e \!\mid\! \x^s,y)
$
is backpropagated to $g$ to update it.

In practice, we utilise a loss that combines gradients from one or more 2D image editors, thus combining their strengths.
Likewise, we can incorporate in this loss additional regularisations to improve the quality of the solution.
Here we consider regularising the depth of the edited shape and appearance of the rendered image. 
We discuss this in detail in the next \Cref{s:2d-editors}.

\paragraph{The choice of $g$.}
Rather than learning the function $g$ from scratch, we note that Shape-E provides a denoising neural network that maps a noised code $\br^s_{\tau} = \alpha_{\tau} \br^s + \sigma_{\tau} \beps$ to an estimate
$
\br^s \approx \hat \br_\text{SE}(\br^s_{\tau}; y, \tau)
$
of the original latent.
We thus set
$
g(\br^e \! \mid \! r^s,y)
= \hat \br_\text{SE}(\br, \tau, y),
$
as an initialisation, where $\br = (\sigma_\tau \br^s + \alpha_\tau \beps, \br^s)$ is obtained by stacking the noised input $\br^s$ with the original latent for a fixed noise level ($\sigma_\tau=0.308$).
This encoding is only useful because the network $g$ is initialized from Shape-E, and it expects a noisy input. In fact, the learned distribution in the original Shap-E is very different from the desired editing distribution.

\input{algo/algo_train}

\subsection{2D editors}%
\label{s:2d-editors}

We consider two types of edits: 
(i) \textbf{global edits} (\eg, \textit{``Make it look like a statue''}), which change the style of the object but preserve its overall structure, and
(ii) \textbf{local edits} (\eg, \textit{``Add a party hat to it''}),
which change the structure of the object locally, but preserve the rest.
To achieve these, we learn our model from a combination of complementary 2D editors and regularisation losses.
For both edit kinds, we adopt a text-guided image-to-image (TI2I) editor for distillation %
and consider further edit-specific priors.

\subsubsection{Global editing}

\paragraph{TI2I loss.}

In order to learn from a pre-trained TI2I model (\eg,~InstructPix2Pix~\cite{brooks2023ip2p}), we obtain the SDS gradient
$
\nabla_{\x_e} \mathcal{L}_\text{SDS-\titi} (\x^e \mid \x^s,y) 
$
from the TI2I denoising network
$
\iptp (\x^e_t; \x^s, y, t).
$
Note that the latter is conditioned on the source image $\x^s$ and the editing instructions $y$.
We also use classifier-free guidance (CFG)~\cite{ho22classifier-free} to enhance the signal of this network for distillation purposes.
Please refer to the \Cref{sec:detailed_implementation} for details.

\paragraph{Depth regularisation for global editing.}

Global edits are expected to change the style of an object, but to retain its overall shape.
We encourage this behaviour via an additional depth regularisation loss:
\begin{equation}
    \mathcal{L}_{\text{reg-global}}(\bd^e, \bd^s) = \mathbb{E}_{\pi} \big[ \lVert \bd^e - \bd^s \rVert ^2 \big],
\end{equation}
where $\bd^e$ and $\bd^s$ are the rendered depth maps from a viewpoint $\pi$ for edited and source objects, respectively.

\paragraph{Overall loss.}
For $\mathcal{L}_{\text{global}}(\x^s, \x^e, \bd^s, \bd^e)$, we use a weighted combination of $\mathcal{L}_\text{SDS-\titi}$ and %
$\mathcal{L}_\text{reg-global}$.

\subsubsection{Local editing}%
\label{s:local-edits}

For local edits, we use $\mathcal{L}_\text{SDS-\titi}$ as before, but also consider additional inductive priors, as follows.

\paragraph{T2I loss.}

Current 2D editors often struggle to edit images locally, %
sometimes failing to apply the edit altogether.
To encourage semantic adherence to the edit instruction, %
we further exploit the semantic priors in a text-to-image (T2I) model, obtaining the SDS gradient
$
\nabla_{\x_e} \mathcal{L}_\text{T2I} (\x^e \!\mid\! y^e)
$
from the denoising network
$
\tti (\x^e_t; y^e, t).
$
Here, the text prompt $y^e$ contains a full description of the edited object (\eg, 
\textit{``A corgi wearing a party hat''}), instead of an instruction based on a reference image.
We use CFG for this gradient as well.

\paragraph{Masked regularisation for local editing.}

To further enhance the locality of the edits, inspired by the cross-attention guidance proposed for controllable generations~\cite{chen2023trainingfree, epstein2023selfguidance}, we extract the cross-attention maps from the pre-trained TI2I model during the SDS loss calculation.
For instance, given a local editing instruction \textit{``Add a party hat to the corgi''}, we compute the cross-attention maps between U-Net features and the specific text embedding for the word \textit{``hat''}.
These maps are then processed to yield a mask $\bm{m}$, which represents an estimation of the editing region.

We can then use the complement of the mask to encourage the appearance of source and edited object to stay constant outside of the edited region:
\begin{align}
\mathcal{L}_{\text{reg-local}}&(\x^s, \x^e, \bd^s, \bd^e, \bm{m})
=
\mathbb{E}_{\pi} \Big[
    (1 - \bm{m}) \nonumber \\ &\odot
    \big(\lambda_{\text{photo}} \lVert
         \x^e - \x^s \lVert^2 
         + \lambda_{\text{depth}} \rVert \bd^e - \bd^s
    \lVert^2 \big)
    \Big] ,
\label{eqn:reg-local}
\end{align}

\noindent where $\lambda_{\text{photo}}$ and $\lambda_{\text{depth}}$ denote corresponding weight factors for the photometric loss 
 $\lVert \x^e - \x^s \rVert^2$ and the depth map differences $\lVert \bd^e - \bd^s \rVert ^2$ between source and edited views. 

\paragraph{Overall loss.}

For $\mathcal{L}_\text{local}(\x^s, \x^e, \bd^s, \bd^e, \bm{m})$, we use a weighted combination of the  $\mathcal{L}_\text{SDS-\titi}$, $\mathcal{L}_\text{SDS-T2I}$ and local regularisation losses $\mathcal{L}_\text{reg-local}$.

%% file: algo/algo_train.tex
\begin{algorithm}[t!]
\small
\caption{\method training}
\label{alg:3DE}

\begin{algorithmic}
\State \textbf{Input:}
$\Theta$: training 3D objects \\
$g$: latent editor initialization \;
$(h,h^*)$: auto-encoder, \\
$\mathcal{L}$: distillation loss \;
$\mathcal{Y}$: instruction set

\State {\bf Output:}
$g$: optimized editor

\While{\text{not converged}}
    \State $\br^s \gets h(\theta^s), \; \theta^s \in \Theta$
    \State $\br^e \gets g(\bm{r^s}, y), \; y \in \mathcal{Y}$ \;
    \State $\triangleright$ Render objects to RGB and depth
    \State $\pi \gets \text{random viewpoint}$
    \State
    $
    (\bm{x}^s, \bm{d}^s) \gets \mathcal{R}(h^*(\br^s), \pi)
    $
    \State
    $
    (\bm{x}^e, \bm{d}^e) \gets \mathcal{R}(h^*(\br^e), \pi) \;
    $
    \State
    Update $g$ using the gradient
    $
    \Delta_g\mathcal{L}(\bm{x}^s, \bm{x}^e, \bm{d}^s, \bm{d}^e)
    $
\EndWhile
\end{algorithmic}
\end{algorithm}

%% file: sec/4_results.tex
\input{tables/clip_comparison}
\begin{figure*}[t!]
    \centering
    \includegraphics[width=\linewidth]{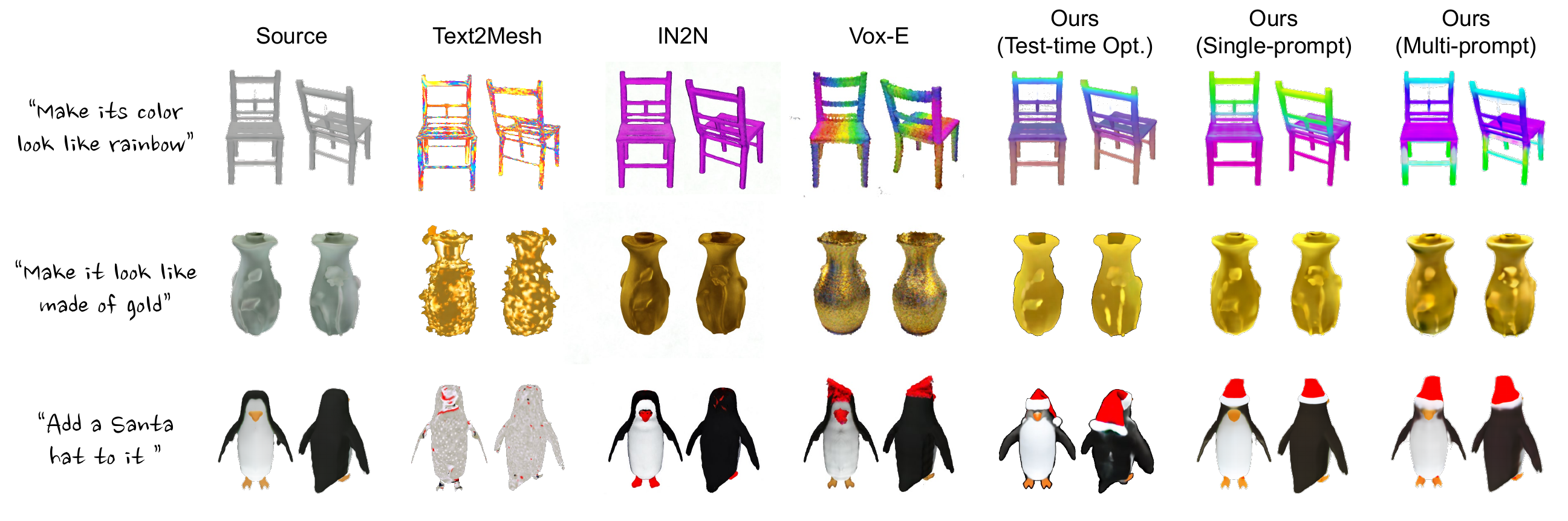}
    \vspace{-0.75cm}
    \caption{\textbf{Qualitative comparison with text-guided 3D editing methods.} Both the single-prompt and multi-prompt versions of our method achieve superior local and global editing results. Our \method can preserve the identity of the original assets, such as the appearance and shape of the \textit{``penguin''}, the fine geometric details of the \textit{``vase''}, and the structure of the \textit{``chair''}.}%
    \label{fig:comparison}
    \vspace{-0.15cm}
\end{figure*}

\section{Experiments}%
\label{sec:exp}

In this section, we provide details of our implementation and the evaluation dataset, %
compare different variants of our approach to state-of-the-art instruction-based editing methods,
and study the effect of the various losses in our approach via ablation.

\subsection{Dataset and implementation details}%
\label{subsec:implementation}

\paragraph{Dataset.}

We construct our 3D object dataset from two sources:
(i) scanned 3D objects from OmniObject3D~\cite{wu2023omniobject3d}, and
(ii) 3D objects generated by Shap-E for specific object categories.
To ensure the high quality of synthetic 3D objects, we apply additional filtering based on their CLIP scores.
The resultant training dataset encompasses approximately $30$ classes, each containing up to $10$ instances. For evaluation, we set up $20$ instance-instruction pairs.
These pairs are composed of $5$ editing instructions ($3$ for global editing and $2$ for local editing) and $15$ high-quality 3D objects \textit{which are not included in the training set} ($8$ objects from Shap-E generation, and $7$ from OmniObject3D).

\paragraph{Evaluation metrics.}

Following common practice~\cite{Sella_2023_ICCV,li2023focaldreamer}, we assess edits by measuring the alignment between generated results and the editing instructions using CLIP similarity (CLIP$_{sim}$) and CLIP directional similarity~\cite{gal2021stylegannada} (CLIP$_{dir}$).
CLIP$_{sim}$ is the cosine similarity between the edited output and the target text prompts.
CLIP$_{dir}$ first calculates the editing directions (\ie, \{target vectors minus source vectors\}) for both rendered images and text descriptions, followed by the evaluation of the cosine similarity between these two directions.
Additionally, to assess structural consistency in \textit{global editing}, we utilise the Structure Distance proposed by~\cite{tumanyan2022splicing}.
This is the cosine similarity between the self-attention maps generated by two images.

\paragraph{Implementation details.}

While training \method, we use IP2P~\cite{brooks2023ip2p} as $\iptp$ for global editing.
For local editing, we employ the Stable Diffusion v1-5 model~\cite{rombach2022high} for $\tti$ and MagicBrush~\cite{Zhang2023MagicBrush} (\ie, a fine-tuned version of IP2P with enhanced editing abilities for object additions) for $\iptp$.
\textit{All} 3D objects used for evaluation, including those in quantitative and qualitative results, are \textit{``unseen''}, \ie, not used to train and thus optimise the editor.
This differs from previous methods that perform test-time optimisation.
Further implementation details are provided in the~\Cref{sec:detailed_implementation}.

\subsection{Comparison to the state of the art}

We compare our method to other text-driven 3D editors such as Instruct-NeRF2NeRF (IN2N)~\cite{instructnerf2023}, Vox-E~\cite{sella2023vox}, and Text2Mesh~\cite{michel2022text2mesh}.
Specifically, Instruct-NeRF2NeRF iteratively updates images rendered from a NeRF with a 2D image editing method (IP2P) and uses the edited images to gradually update the NeRF.
Vox-E optimises a grid-based representation~\cite{karnewar2022relu} by distilling knowledge from a 2D text-to-image model (Stable Diffusion) with volumetric regularisation; a refinement stage is added to achieve localised edits.
Text2Mesh optimises meshes with CLIP similarity between the mesh and the target prompt.
Since different methods receive different input formats (NeRF, mesh, and voxel grid), we provided many ($\sim$ 200) rendered images at 512 $\times$ 512 resolution for initialising their 3D representations.

We consider two variants of \method:
\textbf{(i)~Ours (Single-prompt):} \method trained with a single prompt at a time and multiple classes (this is the default setting for our experiments), and
\textbf{(ii)~Ours (Multi-prompt):} \method trained with multiple prompts and multiple classes.
Finally, we also consider a test-time optimisation baseline (\textbf{Ours (Test-time Optimisation)}), where, instead of training an editor function, the Shape-E latent is optimised directly to minimise the same set of losses.

\paragraph{Quantitative comparison.}

\Cref{tab:clip_comparison} compares methods quantitatively.
Both the single-prompt and multi-prompt variants of our approach are superior to optimisation-based 3D editing methods, despite addressing a harder problem, \ie, the test 3D assets are not seen during training.
The inference of \method is near-instantaneous (within one second) since editing requires only a single forward pass.

\begin{figure*}[t]
    \centering
    \includegraphics[width=0.95\linewidth]{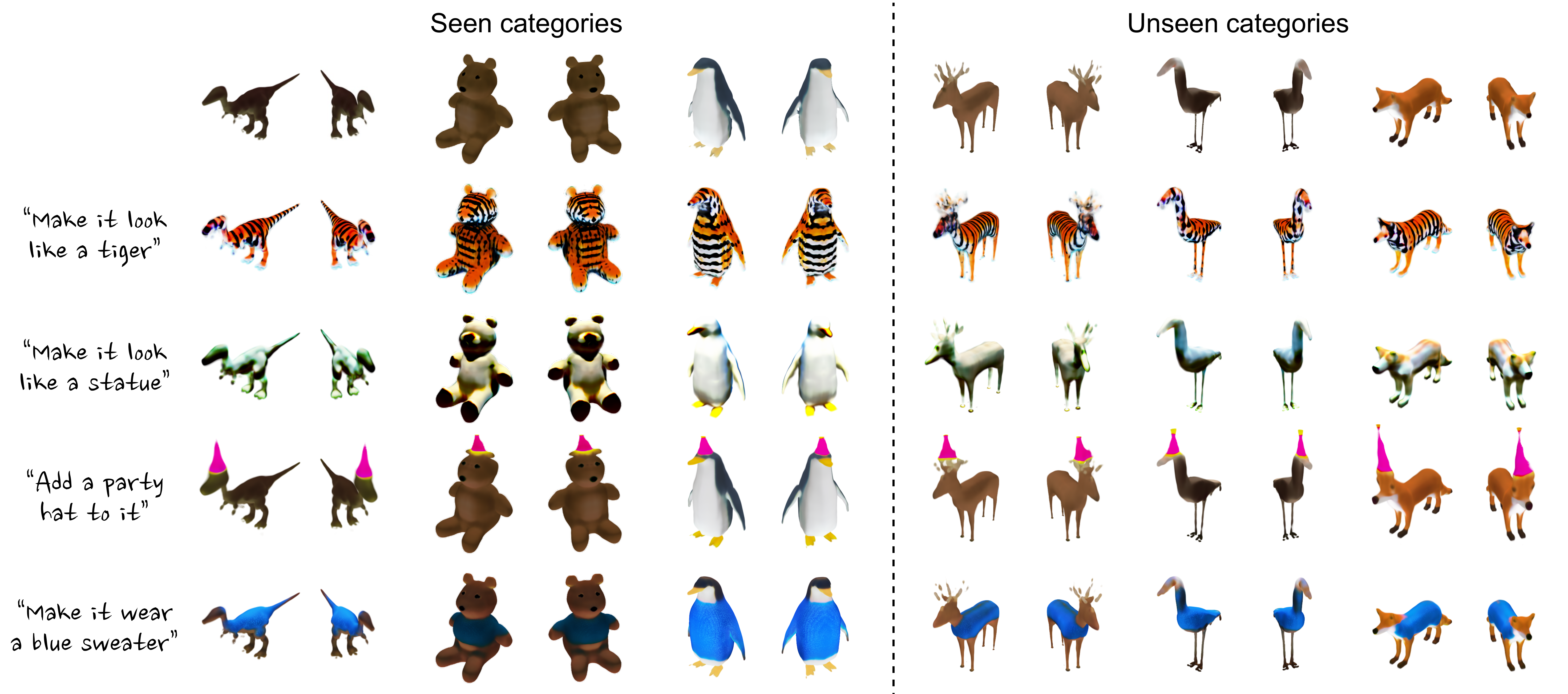}
    \vspace{-0.15cm}
    \caption{\textbf{Generalisation to unseen categories.} ``Seen categories'' refer to object classes included in the training dataset; the specific instances shown were not used for training. ``Unseen categories'' represent the object classes that were never encountered during training.}%
    \label{fig:add-results}
    \vspace{-0.2cm}
\end{figure*}

\paragraph{Qualitative comparison.}

\Cref{fig:comparison} compares methods qualitatively.
All prior works struggle with global edits.
Text2Mesh results in noisy outputs and structural changes.
IN2N is able to preserve the shape and identity of the original objects but fails to converge for some prompts, such as \textit{``Make its color
look like rainbow''}.
The reason is that edited images produced by IP2P share almost no consistency under this prompt, which cannot be integrated coherently into 3D.
On the other hand, Vox-E successfully changes the appearance of the objects, but due to distillation from a T2I model rather than a TI2I model, it fails to preserve the geometry.

When local edits are desired, such as \textit{``Add a Santa hat to it''} (\Cref{fig:comparison}, bottom row), Text2Mesh and IN2N do not produce meaningful changes.
Text2Mesh mainly changes the texture, and IN2N ignores the instruction entirely.
This can be attributed to the inability of their underlying 2D models to add or remove objects.
Vox-E adds the hat to the penguin, but other regions (\eg, nose) also change unintentionally, despite their spatial refinement stage.

The combination of training objectives in our approach leverages the complementary aspects of different 2D diffusion priors, overcoming these problems even while using feed-forward prediction.
Furthermore, the learned editor also improves over test-time optimisation results with the same prompt and optimisation objectives.
We hypothesise that this is because learning an editor can regularise the editing process too.
Finally, while a single-prompt editor achieves the best results, we show that it is possible to train an editor with multiple prompts (last column) without compromising fidelity or structure.

\Cref{fig:add-results} provides additional results for various instructions, each associated with a single-prompt editor. Our trained editors are capable of performing consistent edits across diverse objects, and, importantly, generalise to \emph{unseen categories} not included in the training dataset.

\begin{figure*}[t]
    \centering
    \includegraphics[width=0.97\linewidth]{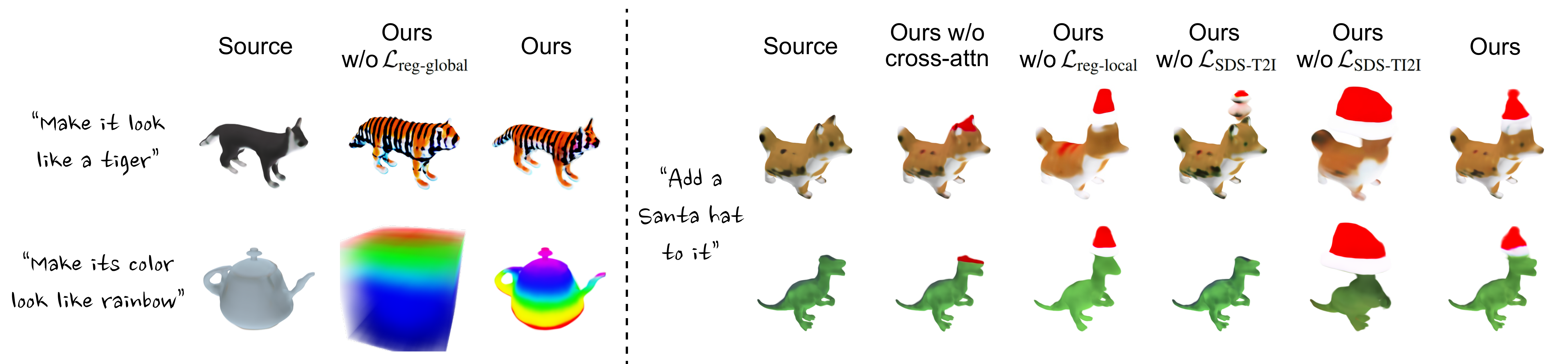}
    \vspace{-0.1cm}
    \caption{\textbf{Qualitative ablation results,} where the left and right parts correspond to global and local editing, respectively.}%
    \label{fig:ablation}
    \vspace{-0.1cm}
\end{figure*}

\subsection{Ablation study}%
\label{subsec:ablation}

\input{tables/ablation_table}

\paragraph{Quantitative analysis.}

\Cref{tab:abla-global} presents the quantitative results for global editing, where the omission of depth regularisation %
leads to a noticeable degradation in performance, reflected by high Structure Dist.
Likewise, the removal of loss components for local editing impairs the model to varying extents (\Cref{tab:abla-local}), which we analyse next.

\paragraph{Qualitative analysis.}

In \Cref{fig:ablation}, we illustrate the effect of the different model components.
For global editing, eliminating the depth regularisation term (\ie, Ours w/o $\mathcal{L}_{\text{reg-global}}$) can lead to significant alterations of the source shape. %
For local editing, we observe the following: %
(i) the \textbf{cross-attn masks} specify the editable region where regularisation is not applied.
If such a region is not defined, the depth and photometric regularisers would be applied to the whole object, thereby forbidding the formation of local shapes (in this case, the \textit{Santa hat}); 
(ii) the \textbf{regularisation loss} ($\mathcal{L}_{\text{reg-local}}$) helps the model to maintain the object's identity (both appearance and shape);
(iii) the \textbf{T2I loss} ($\mathcal{L}_{\text{SDS-T2I}}$) significantly improves the quality of local editing. 
When omitted (\ie, Ours w/o $\mathcal{L}_{\text{SDS-T2I}}$), only the TI2I prior is used, which struggles with localised edits (same issues that \cite{instructnerf2023, kamata2023instruct} exhibit); 
(iv) the \textbf{TI2I loss} ($\mathcal{L}_{\text{SDS-TI2I}}$) uses source images as references, which greatly helps with understanding the layout of edits.
Thus, Ours w/o $\mathcal{L}_{\text{SDS-TI2I}}$ leads to spatial inaccuracy in editing (same as \cite{Sella_2023_ICCV}).

\begin{figure}[t]
    \centering
    \includegraphics[width=\linewidth]{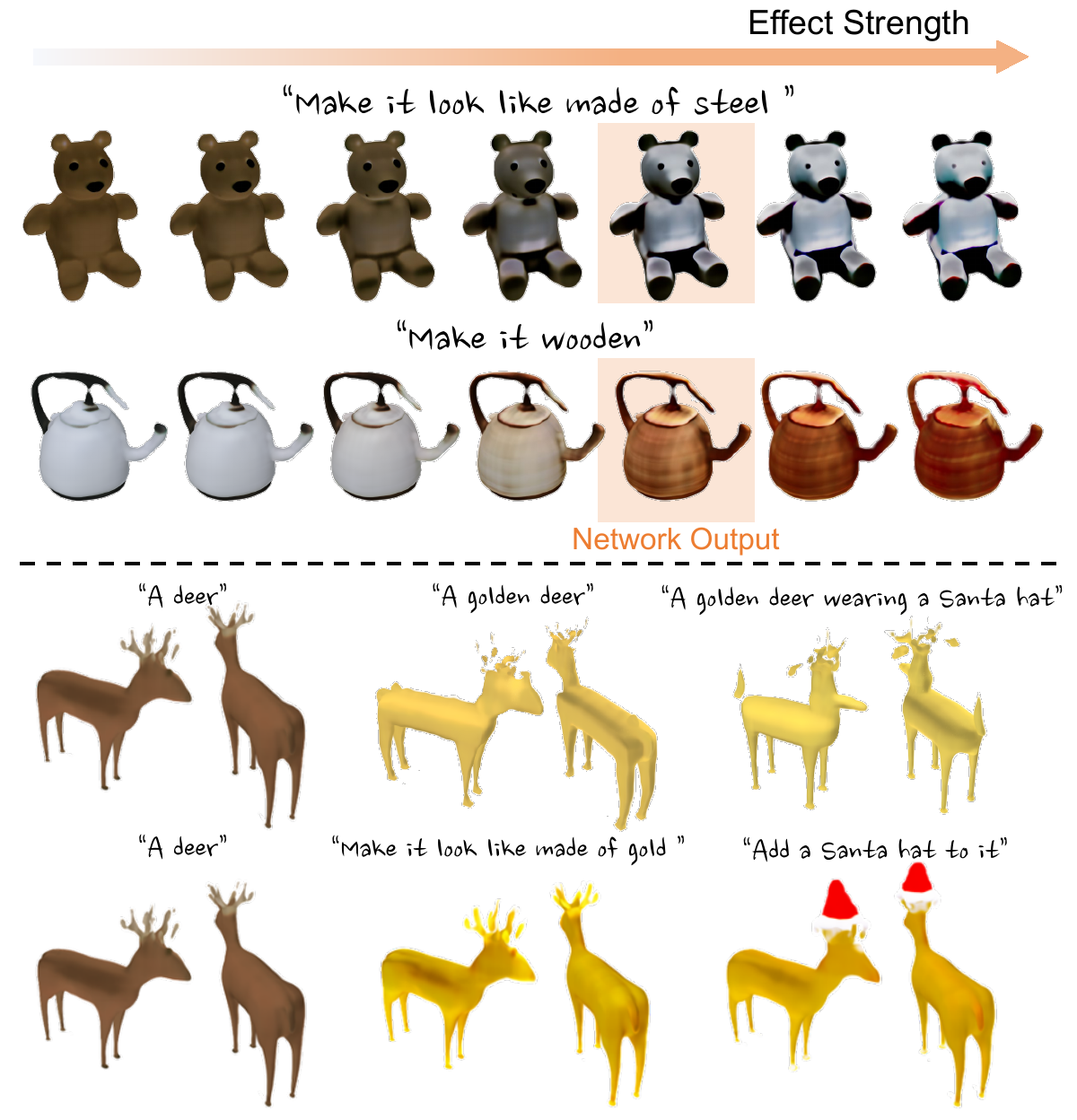}
    \vspace{-0.65cm}
    \caption{\textbf{Top}: the strength of the editing effects can be controlled via linear interpolation and extrapolation in latent space. \textbf{Bottom}: the examples in the first row are directly generated by Shap-E and the second row is generated by progressively adding multiple effects to the unseen category \textit{``deer''}.}%
    \label{fig:progressive}
    \vspace{-0.4cm}
\end{figure}

\subsection{Discussion}

\begin{figure}[t]
    \centering
    \hspace{0.05cm}
    \includegraphics[width=\linewidth]{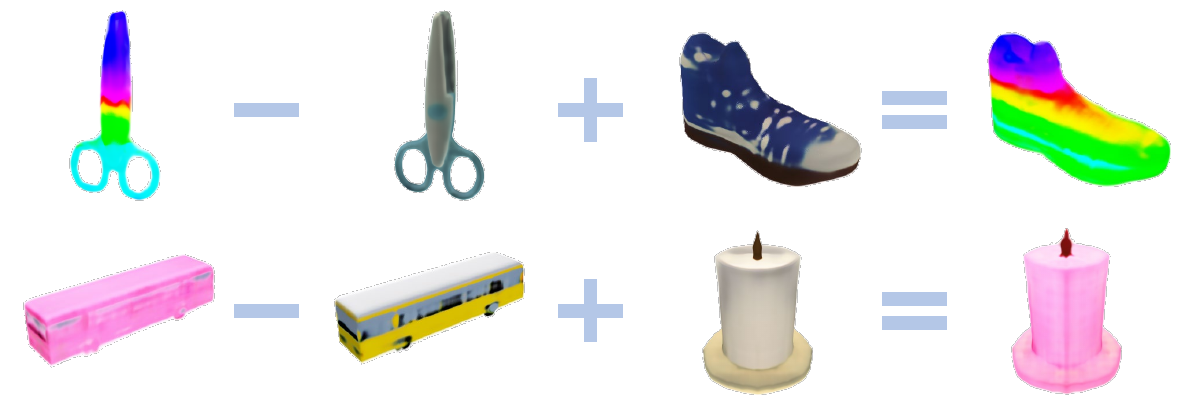}
    \vspace{-0.5cm}
    \caption{\textbf{Unified editing vector.} The editing effects can be transferred via simple vector arithmetic operations in latent space.}%
    \label{fig:arithmetic}
    \vspace{-0.3cm}
\end{figure}

In \Cref{fig:progressive} (top), we observe that the latent space of Shap-E is partially linear.
After training the editor to produce the desired effects, we can further control the strength of the effects.
This could be done by scaling to residual of updated latent and source latent by a factor $\eta$.
The editor's output corresponds to $\eta=1$.
Increasing (decreasing) $\eta$ weakens (strengthens) the effects.
In \Cref{fig:progressive} (bottom), we show that edits can be accumulated progressively until the desired effect is achieved.
Furthermore, as noted in~\cite{jun2023shap} and shown in the figure, Shap-E (the first row of the bottom part) itself fails at compositional object generation, but our approach can largely remedy that by decomposing complex prompts into a series of edits.
Finally, in \Cref{fig:arithmetic}, we also show that some of the edits, once expressed in latent space, are quite linear.
By this, we mean that we can find a single vector for effects like \textit{``Make its color look like rainbow''} or \textit{``Turn it into pink''} that can be used to edit any object by mere addition regardless of the input latent.
This is a strong indication that the latent space is well structured and useful for semantic tasks like editing.

\paragraph{Limitations.}

Our work is based on the latent space of Shap-E and pre-trained 2D editors, which pose an upper bound on quality and performance.
Furthermore, while we show that we can learn a latent editor that understands multiple instructions, we could not yet achieve a fully open-ended editor.
We conjecture that this might require training at a much larger scale than we can afford (\ie, hundreds of GPUs \textit{vs.}\@ a handful).

%% file: tables/clip_comparison.tex
\begin{table*}[t]
  \centering
  \footnotesize
    \begin{tabular}{lc cc ccc c}
        \toprule
        \multirow{2}[2]*{\textbf{Model}} & \multirow{2}[2]{*}{\shortstack{\textbf{Per-instance} \\ \textbf{optimization}}} & \multicolumn{2}{c}{\textbf{Local editing}}  & \multicolumn{3}{c}{\textbf{Global editing}} & \multirow{2}[2]{*}{\textbf{Inference time} $\downarrow$}  \\
        \cmidrule(lr){3-4}         \cmidrule(lr){5-7}
     &  & CLIP$_{sim}\uparrow$  & CLIP$_{dir}\uparrow$  & CLIP$_{sim}\uparrow$  & CLIP$_{dir}\uparrow$ & Structure Dist. $\downarrow$ & \\
        \midrule
        Text2Mesh \cite{michel2022text2mesh} & \cmark & 0.239 & 0.058 & 0.248 & 0.057 & 0.073 & $\sim$ 14 min \\
        Instruct-NeRF2NeRF \cite{instructnerf2023} & \cmark &  0.253 & 0.051 & 0.239 & 0.057 & 0.095 & $\sim$ 36 min \\
        Vox-E \cite{Sella_2023_ICCV}  & \cmark &   0.277  &  0.075   & 0.271 &  0.066 & 0.026 & $\sim$ 40 min (+ 13 min) \\ %
        \midrule
        Ours (Test-time Opt.) & \cmark & 0.290  & 0.087 & 0.268 &\textbf{0.072}  & 0.013 & $\sim$ 19 min \\
        Ours (Single-prompt)  & \xmark & \textbf{0.292} & \textbf{0.097} & \textbf{0.272} & 0.070  &  \textbf{0.008}  & $\sim$ \textbf{1 sec}\\
        Ours (Multi-prompt)  & \xmark & 0.279 & 0.085 & 0.255 & 0.062  & 0.009 & $\sim$ \textbf{1 sec} \\
        \bottomrule
    \end{tabular}
    \vspace{-0.15cm}
    \caption{\textbf{Quantitative comparison of our \method with other per-instance editing methods}. The measured inference time excludes both the rendering process and the encoding of 3D representations. The time inside the bracket indicates the extra time required by Vox-E for its refinement step in local editing. Our method achieves superior results within one second on the evaluation dataset. }
    \label{tab:clip_comparison}
    \vspace{-0.2cm}
\end{table*}

%% file: tables/ablation_table.tex
\begin{table}[t]
    \centering
    \footnotesize
    \begin{subtable}{\linewidth}
    \centering
    \setlength\tabcolsep{4pt}
    \begin{tabular}{lccc}
        \toprule 
         \textbf{Model} & CLIP$_{sim}$ $\uparrow$ & CLIP$_{dir}$ $\uparrow$ & Structure Dist. $\downarrow$ \\
        \midrule
        Ours w/o $\mathcal{L}_{\text{reg-global}}$  & 0.218 & 0.058 & 0.138 \\
        \midrule
         Ours & \textbf{0.272} & \textbf{0.070} & \textbf{0.008} \\
        \bottomrule
    \end{tabular}
    \caption{\footnotesize Ablation study for global editing.}
    \label{tab:abla-global}
    \end{subtable}%
    \\
    \par\vspace{0.1cm}
    \begin{subtable}{\linewidth}
    \centering
    \setlength\tabcolsep{10pt}
    \begin{tabular}{lcc}
        \toprule 
         \textbf{Model} & CLIP$_{sim}$  $\uparrow$ & CLIP$_{dir}$  $\uparrow$  \\
        \midrule
        Ours w/o cross-attn masks & 0.261 & 0.064  \\
        Ours w/o $\mathcal{L}_{\text{reg-local}}$  & 0.282 & 0.092 \\
        Ours w/o $\mathcal{L}_{\text{SDS-T2I}}$ & 0.263 & 0.067\\
        Ours w/o $\mathcal{L}_{\text{SDS-TI2I}}$ & 0.278 & 0.096\\
        \midrule
        Ours  & \textbf{0.292} & \textbf{0.097}  \\
        \bottomrule
    \end{tabular}
    \caption{\footnotesize Ablation study for local editing.}
    \label{tab:abla-local}
    \end{subtable}%
    \vspace{-0.2cm}
    \caption{\textbf{Quantitative ablation study on loss components.}}
    \vspace{-0.5cm}
\end{table}

%% file: sec/5_conclusion.tex
\section{Conclusion}%
\label{sec:conclusion}

We have introduced \method, a universal editor for different 3D objects that operates efficiently in latent space.
It eschews costly test-time optimisation and runs in a feed-forward fashion within one second for any object.
\method is trained from multiple 2D diffusion priors and thus combines their strengths, achieving compelling results for both global and local edits, even when compared to slower optimisation-based 3D editors.

%% file: sec/supp.tex
\section*{Appendix}
\label{appendix}
\appendix
\section{Overview}
This Appendix contains the following parts:
\begin{itemize}
    \item \textbf{Implementation details (\Cref{sec:detailed_implementation}).} We provide full details regarding the dataset formulation and experimental settings.

    \item \textbf{Additional results (\Cref{sec:addtion_result}).} We provide additional examples of our method, including additional global and local prompts, and we further demonstrate the potential of our multi-prompt editor to handle a large number of prompts.
    \item \textbf{Additional ablation study (\Cref{sec:ablation2}).} We provide additional ablation studies of \method on the initialisation method, the choice of $\sigma_{\tau}$, and the attention maps used to guide the regularisation loss for local editing.
    \item \textbf{Extended discussion on prior methods (\Cref{sec:ext_dicussion}).} We discuss the difference between our \method and other related 3D editing methods.
    \item \textbf{Failure cases (\Cref{sec:failure_case}).} We provide failure cases of our method qualitatively.
\end{itemize}

\section{Implementation details}
\label{sec:detailed_implementation}

\subsection{Dataset formulation}
In this section, we provide more details regarding the construction of the training and evaluation datasets.

\paragraph{Training dataset.} The training dataset specifies a set of 3D objects used to train different instructions. There are in total $33$ object classes, each containing up to $10$ instances.

Specifically, we use Shap-E-generated objects spanning $20$ object classes: 
\textit{apple, banana, candle, cat, chair, corgi, dinosaur, doctor, duck, guitar, horse, microphone, penguin, pineapple, policeman, robot, teapot, teddy bear, toy plane, vase}

In addition, we use 3D objects from OmniObject3D~\cite{wu2023omniobject3d} spanning $21$ object classes:  
\textit{bear, cat, cow, dinosaur, dog, duck, elephant, giraffe, guitar, hippo, mouse, panda, pineapple, rabbit, rhino, scissor, teapot, teddy bear, toy plane, vase, zebra}

Note that, we manually filter out the invalid instruction-instance pairs during training. For example, we consider it unreasonable to \textit{``add a Santa hat''} to \textit{``chairs''}, thereby discarding such pairs. Consequently, we obtain a set of valid object classes for each editing instruction during training, as summarised in~\Cref{suptab: traindataset}.

\paragraph{Evaluation dataset.} The evaluation dataset consists of $20$ high-quality instance-instruction pairs ($12$ global editing pairs and $8$ local editing pairs), with the details listed in Table~\ref{suptab: evaldataset}. In summary, there are $3$ and $2$ editing instructions for global and local editing, respectively, with $8$ Shap-E generated objects and $7$ instances sourced from OmniObject3D. Note that \textit{none} of the instances in the evaluation dataset are utilised for training purposes.

\newcommand{\centered}[1]{\begin{tabular}{c} #1 \end{tabular}}
\newcommand{\toleft}[1]{\begin{tabular}{l} #1 \end{tabular}}
\begin{table*}[t]
    \centering
    \footnotesize
    \setlength\tabcolsep{4pt}
    \begin{tabular}{ccc}
    \textbf{Editing type} & \textbf{Instruction} & \textbf{Object class} \\
     \toprule
      \centered{Global} & \centered{\textit{``Make it look like made of gold''}} & \centered{\textit{apple, banana, candle, cat, chair, corgi, dinosaur, doctor, duck, guitar, horse,} \\ \textit{microphone, penguin, pineapple, policeman, robot, teapot, teddy bear, toy plane, vase;} \\ \textit{\textbf{bear, cat, cow, dinosaur, dog, duck, elephant, giraffe, guitar, hippo, mouse,}} \\ \textit{\textbf{panda, pineapple, rabbit, rhino,  scissor, teapot, teddy bear, toy plane, vase, zebra}} } \\
     \midrule
     \centered{Global} & \centered{\textit{``Make it look like a tiger''}} & \centered{\textit{cat, corgi, dinosaur, duck, horse, penguin, teddy bear;} \textit{\textbf{bear, cat, cow, dinosaur,}} \\ \textit{\textbf{dog, duck, elephant, giraffe, hippo, mouse, panda, rabbit, rhino, teddy bear, zebra}} } \\
      \midrule
     \centered{Global} & \centered{\textit{``Make its color look like rainbow''}} & \centered{\textit{apple, banana, candle, cat, chair, corgi, dinosaur, doctor, duck, guitar, horse,} \\ \textit{microphone, penguin, pineapple, policeman, robot, teapot, teddy bear, toy plane, vase;} \\ \textit{\textbf{bear, cat, cow, dinosaur, dog, duck, elephant, giraffe, guitar, hippo, mouse,}} \\ \textit{\textbf{panda, pineapple, rabbit, rhino,  scissor, teapot, teddy bear, toy plane, vase, zebra}} } \\
      \midrule
     \centered{Local} & \centered{\textit{``Add a Santa hat to it''}} & \centered{\textit{cat, corgi, dinosaur, doctor, duck, horse, penguin, policeman, teddy bear;} \textit{\textbf{bear, cat, cow,}} \\ \textit{\textbf{dinosaur, dog, duck, elephant, giraffe, hippo, mouse, panda, rabbit, rhino, teddy bear, zebra}}} \\ 
      \midrule
     \centered{Local} & \centered{\textit{``Make it wear a blue sweater''}} & \centered{\textit{cat, corgi, dinosaur, doctor, duck, horse, penguin, policeman, teddy bear;} \textit{\textbf{bear, cat, cow,}} \\ \textit{\textbf{dinosaur, dog, duck, elephant, giraffe, hippo, mouse, panda, rabbit, rhino, teddy bear, zebra}}} \\ 
     \bottomrule
    \end{tabular}
    \vspace{-0.15cm}
    \caption{\textbf{Training dataset formulation.} The object classes in \textbf{\textit{bold}} are sourced from OmniObject3D%
    , whereas the remaining classes are generated from text prompts using Shap-E.}
    \label{suptab: traindataset}
    \vspace{0.2cm}
\end{table*}

\begin{table*}[t]
    \centering
    \footnotesize
    \begin{tabular}{ccc}
    \textbf{Editing type} & \textbf{Instruction} & \textbf{Instance} \\
     \toprule
      &  \textit{``Make it look like made of gold''} & \textit{``A bird'', \textbf{``An apple''},  \textbf{``A scissor''}, \textbf{``A vase''}} \\
      Global&  \textit{``Make it look like a tiger''} & \textit{``A cat'', ``A corgi'', \textbf{``A dinosaur''}, \textbf{``A zebra''}}  \\
     &  \textit{``Make its color look like rainbow''} & \textit{``A chair'', ``A teapot'', \textbf{``A guitar''}, \textbf{``A pineapple''}}  \\
     \midrule
     \multirow{2}{*}{Local}   &  \textit{``Add a Santa hat to it''} & \textit{``A corgi'', ``A penguin'', ``A robot'', \textbf{``A dinosaur''}, \textbf{``A teddy bear''}}  \\
      &  \textit{``Make it wear a blue sweater''} & \textit{``A corgi'', ``A penguin'', \textbf{``A teddy bear''}} \\
     \bottomrule
    \end{tabular}
    \vspace{-0.1cm}
    \caption{\textbf{Evaluation dataset formulation.} The instances in \textbf{\textit{bold}} are sourced from OmniObject3D%
    , whereas the remaining instances are generated from text prompts using Shap-E. The specific object instances used for evaluation are not seen during training.}
    \label{suptab: evaldataset}
    \vspace{0.2cm}
\end{table*}

\subsection{Experimental details}
\paragraph{Shap-E settings.}
The encoder $h$ takes as input an RGB point cloud ($16384$ points) and different views ($20$) of the 3D asset from random camera angles at $256 \times 256$ resolution. The outputs of the encoder are latents with shape $1024\times1024$.

The decoder outputs the parameters of a neural field represented as a $6$-layer MLP. The weights of the first four layers are linear transformations of the latent, while the weights of the last two layers are fixed.
The output feature vector computed through the MLP is then mapped to the neural field's density and RGB values (or alternatively, SDF and texture color) using different heads.

Finally, Shap-E uses a generative latent-space model for which it employs a transformer-based diffusion architecture akin to Point-E \cite{nichol2022point}, with latent dimensions of $1024 \times 1024$.
It offers two pre-trained conditional diffusion models: image-conditional and text-conditional. The image-conditional approach, paralleling Point-E, augments the transformer context with a $256$-token CLIP embedding. The text-conditional model introduces a single token to the transformer context. We use the text-conditional model in our paper.

\paragraph{SDS with classifier guidance.}
During the Score Distillation Sampling (SDS) process, we adopt the classifier-free guidance~\cite{ho22classifier-free} to enhance the signal of each underlying 2D model for distillation purposes. Specifically, for the text-guided image-to-image (TI2I) SDS, we define:
\begin{align}
\iptp^* &(\x^e_t; \x^s, y, t)
=
\iptp (\x^e_t; \bm{\varnothing}, \varnothing, t)
\nonumber \\
& + \gamma_I \cdot \big(
\iptp (\x^e_t; \x^s, \varnothing, t) -
\iptp (\x^e_t; \bm{\varnothing}, \varnothing, t)
\big)
\nonumber \\
& + \gamma_T \cdot \big(
\iptp (\x^e_t; \x^s, y, t) -
\iptp (\x^e_t; \x^s, \varnothing, t)
\big),
\end{align}
where $\gamma_I$ and $\gamma_T$ correspond to image and text guidance scales, respectively. 
Then: 
\begin{equation}
\resizebox{0.9\linewidth}{!}{%
$ \nabla_{\x_e} \mathcal{L}_\text{SDS-\titi} (\x^e \mid \x^s,y) = 
\mathbb{E}_{t, \beps}
\Big[
\iptp^* (\x^e_t; \x^s, y, t)  - \beps
\Big]$%
}
\end{equation}
\noindent
Similarly, for text-to-image (T2I) SDS, 
\begin{multline}
\tti^* (\x^e_t; y^e, t)
=
\tti (\x^e_t; \varnothing, t) \\
+ \gamma'_T \cdot
\big(
\tti (\x^e_t; y^e, t) -
\tti (\x^e_t; \varnothing, t)
\big),
\end{multline}
where $\gamma'_T$ denotes the text guidance scale, and 
\begin{equation}
\nabla_{\x_e} \mathcal{L}_\text{SDS-T2I} (\x^e \!\mid\! y^e) = 
\mathbb{E}_{t, \beps}
\Big[
\tti^* (\x^e_t; y^e, t)  - \beps
\Big]
\end{equation}

For global editing, where only TI2I SDS is applied, we consider a default setting of guidance scales $(\gamma_I, \gamma_T) = (2.5, 50)$. For local editing, we adopt the guidance scales $(\gamma_I, \gamma_T,  \gamma'_T) = (2.5, 7.5, 50)$.

\paragraph{Loss configuration.}
In terms of the overall loss for global editing, we consider a weighted combination of \titi{ }and global regularisation losses,
\begin{align}
    \mathcal{L}_{\text{global}}(\x^s, \x^e, \bd^s, \bd^e) & = 
    \lambda_\text{\titi} \cdot \mathcal{L}_\text{SDS-\titi} (\x^e \!\mid\! \x^s, y) \nonumber \\ 
    & + \lambda_\text{reg-global}  \cdot \mathcal{L}_\text{reg-global} (\bd^e, \bd^s),
\end{align}
with loss scales indicated by $\lambda_\text{\titi}$ and $ \lambda_\text{reg-global}$, respectively.

For local editing, we use a weighted combination of the \titi, T2I, and local regularisation losses:
\begin{align}
    \mathcal{L}_{\text{local}}(\x^s, \x^e, \bd^s, \bd^e, \bm{m} & )  =    \lambda_\text{\titi}  \cdot \mathcal{L}_\text{SDS-\titi} (\x^e \!\mid\! \x^s, y)  \nonumber \\ 
    + & \lambda_\text{T2I} \cdot \mathcal{L}_\text{SDS-T2I} (\x^e \!\mid\! y^e)  \nonumber\\ 
    + & \mathcal{L}_\text{reg-local} (\x^s, \x^e, \bd^s, \bd^e, \bm{m}),
\end{align}
where $\lambda_\text{\titi}$ and $\lambda_\text{T2I}$ denote corresponding loss scales, and $\mathcal{L}_\text{reg-local} (\x^s, \x^e, \bd^s, \bd^e, \bm{m})$ is defined by ~\cref{eqn:reg-local}.

\paragraph{Estimation of local editing region.} 
An estimate of local editing regions can be obtained by extracting the cross-attention maps from pre-trained 2D models (\ie, MagicBrush). 
Specifically, given an example editing prompt \textit{``Add a Santa hat to it''}, we first calculate the cross-attention maps between the image features and the word token \textit{``hat''}. 
We then average all cross-attention maps corresponding to feature resolution $32 \times 32$ at a particular timestep $t=600$. 
The averaged map undergoes a series of post-processing steps, including (i) bilinear upsampling to a higher resolution at $128 \times 128$; (ii) hard thresholding at $0.5$; (iii) spatial dilation by $10$ pixels; (iv) Gaussian blurring by $5$ pixels. 
The final mask $\bm{m}$ is then adopted as an approximation of the editable region, and used in ~\cref{eqn:reg-local}.

\paragraph{Model settings.} As mentioned previously, we consider two variants in our method, namely \textbf{Ours (Single-prompt)} and \textbf{Ours (Multi-prompt)}. 
We train both methods on objects from the entire training dataset, to ensure their applicability across multiple instances of various categories.

In terms of the editing instructions, \textbf{Ours (Single-prompt)} is considered as our default model and designed to handle \textit{one} prompt at one time. Consequently, it requires $5$ independent models for each of the editing instructions in the evaluation dataset. In contrast, \textbf{Ours (Multi-prompt)} is trained on a combination of editing instructions and is capable of performing different edits according to the input text prompt. We train this multi-prompt model on all $5$ instructions from the evaluation dataset simultaneously.

\paragraph{Architecture details.} 
For \method's architecture, we use a similar network architecture as the text-conditional diffusion model in Shap-E \cite{jun2023shap}, a transformer-based network. The original Shap-E text-to-3D network takes a noisy latent $\sigma_\tau \br^s + \alpha_\tau \beps$ as input and directly predicts the original clean latent $\br^s$.
Instead, the goal of our editor is to transform the original latent $\br^s$ to an edited one.
Therefore, to support $(\sigma_\tau \br^s + \alpha_\tau \beps, \br^s)$ as our input, we add additional input channels to the first linear projection layer.
All weights are initialised by the weights of the pre-trained Shap-E text-to-3D diffusion model, while the weights that apply to the additional input channels are initialized to zeros following a similar setting to \cite{brooks2023ip2p}.

\paragraph{Rendering details.} 
During the training phase, camera positions are randomly sampled using a circular track. This track has a radius of $4$ units and a constant elevation angle of $30^\circ$. The azimuth angle varies within the range of [$-180^\circ$, $180^\circ$]. For loss computation, images of the source NeRF and edited NeRF are rendered at a resolution of $128 \times 128$. 

\paragraph{Training details.}
We adopt a constant learning rate of $1e^{-4}$, utilising the Adam optimiser with $\beta = (0.9, 0.999)$, a weight decay of $1e^{-2}$, and $\epsilon = 1e^{-8}$. The batch size is $64$. We train our single-prompt and multi-prompt models for $150$ and $500$ epochs, respectively. We use the same timestep for $\iptp (\x^e_t; \x^s, y, t)$ and $\tti (\x^e_t; y^e, t)$, which is randomly sampled from $[0.02, 0.98]$ following the setting in \cite{poole2022dreamfusion}.
We also adopt an annealing schedule for the max timestep.
After $100$ and $300$ epochs for the single- and multi-prompt versions, respectively, the max timestep of $t$ decreases with a ratio of $0.8$ for every $10$ epochs and $50$ epochs.
The annealing scheduler helps the model capture more details in the late training.

We set $\lambda_\text{T2I} = \lambda_\text{TI2I} = 1$, and the regularisation terms $\lambda_{\text{photo}}=1.25$ and $\lambda_{\text{depth}}=0.8$ for local editing and $\lambda_{\text{reg-global}}=5$ for global editing in order to better preserve structure. We also employ a linear warmup schedule for the photometric loss in the early epochs. This helps the model to first focus on generating correct semantics, such as \textit{``a penguin wearing a Santa hat''}, and then gradually reconstructing the appearance and shape of the original object (\eg, the \textit{``penguin''}) with the help of masked loss, \ie recovering the identity of the original object.
The training of each single-prompt model takes approximately $10$ GPU hours, and the multi-prompt model takes $30$ GPU hours, on the NVIDIA RTX A6000. 

\paragraph{Evaluation details.}
During evaluation, to compute the CLIP metrics and the Structure Distance, we uniformly sample $20$ viewpoints following the same recipe as in the training phase. All rendered images are resized to $256 \times 256$ to ensure a fair comparison across different methods.

\section{Additional Results}
\label{sec:addtion_result}

\begin{figure*}
\centering
\includegraphics[width=\linewidth]{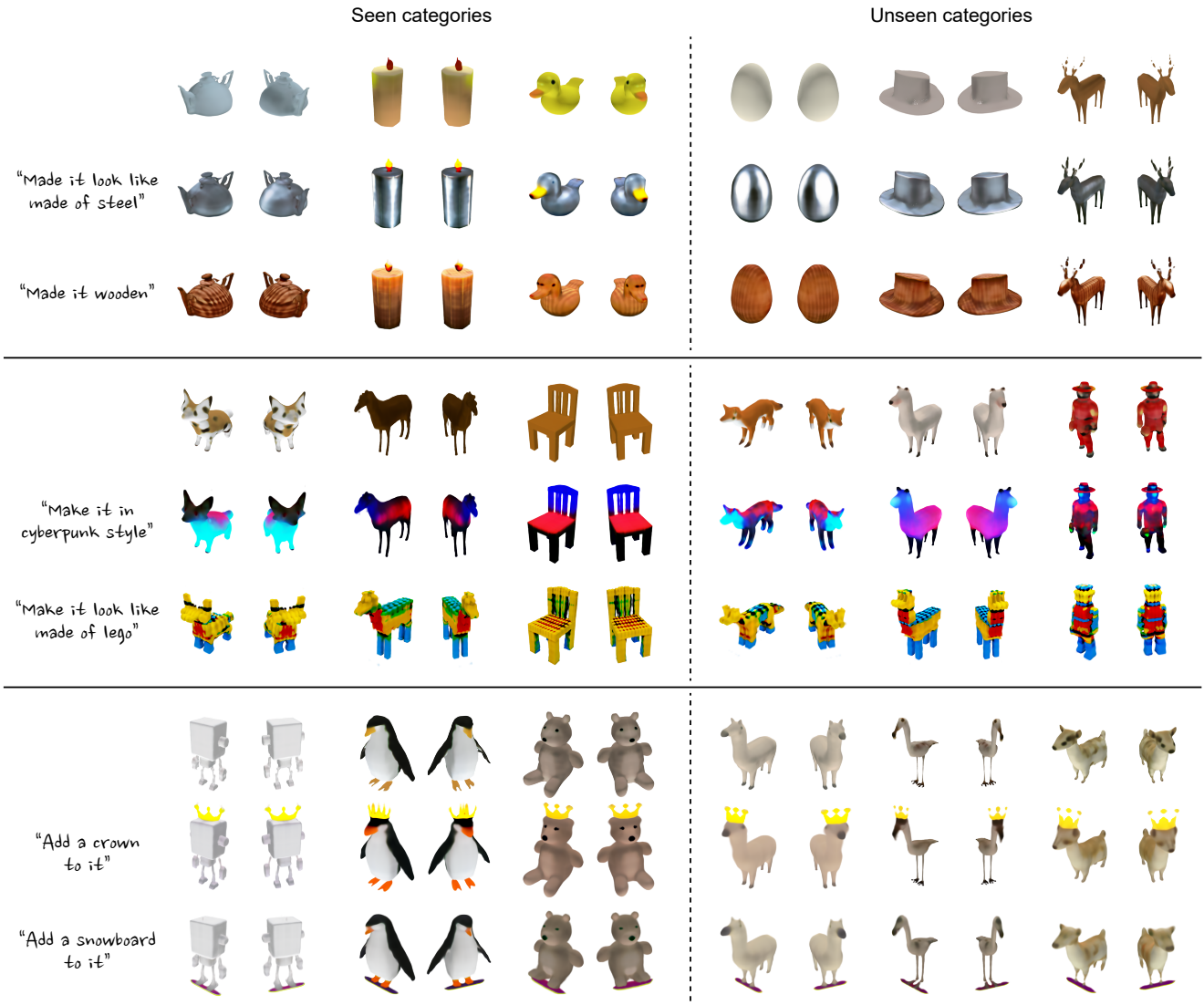}
\vspace{-0.6cm}
\caption{\textbf{Additional visualisations.} We apply different editing instructions (including both global and local edits) across various instances, also demonstrating the generalisability of our method to multiple unseen 
 categories.}%
\label{fig:additional-sup}
\vspace{-0.cm}
\end{figure*}

\paragraph{Additional visualisations.} \Cref{fig:additional-sup} provides additional visualised results for our \method, with each editing prompt associated with a distinct model, \ie, Ours (Single-prompt). It is evident that our method is capable of performing accurate edits across diverse object classes and demonstrates reasonable generalisation to unseen categories.

\begin{figure}
\centering
\includegraphics[width=\linewidth]{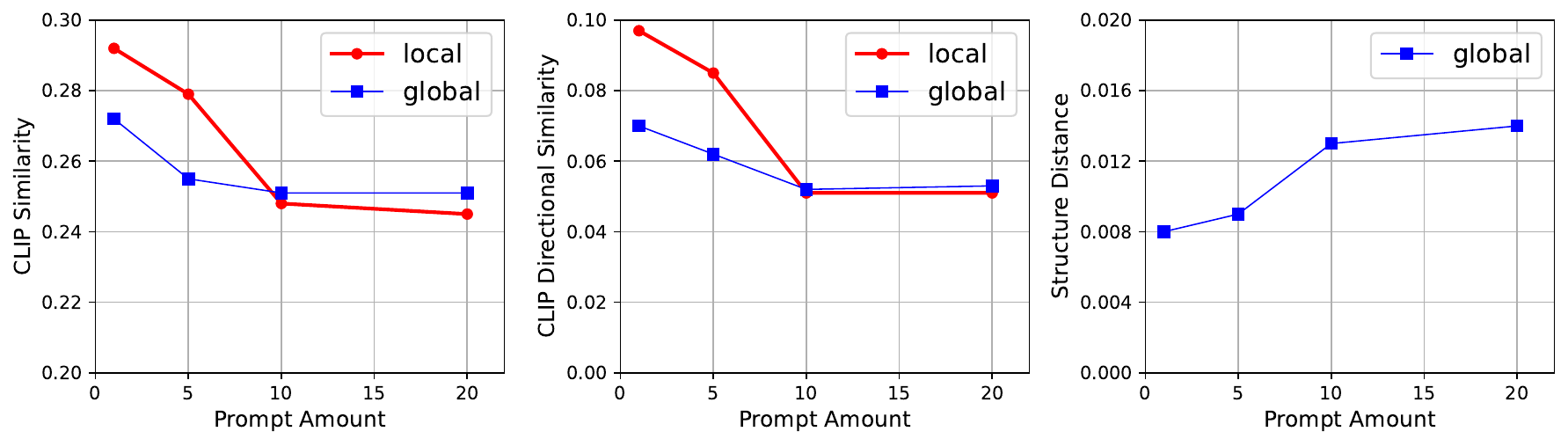}
\vspace{-0.6cm}
\caption{\textbf{Scale up the amount of prompts.} We explore the possibility of learning a single editor function with tens of prompts. As the amount of prompts increases, the CLIP similarly and CLIP directional similarity scores decrease. However, both scores reach a plateau when more prompts are introduced.}%
\label{fig:scale}
\vspace{-0.cm}
\end{figure}

\paragraph{Scaling up to more prompts.}
We further explore the possibility of learning more prompts within one editor model.
We first train a $10$-prompt model using the five instructions included in the original dataset 
plus extra five prompts: \textit{``Make it look like a statue''}, \textit{``Make it look like made of steel''}, \textit{``Make it look like made of lego''}, \textit{``Add a party hat to it''}, \textit{``Add rollerskates to it''}.
We also expand the instructions to train a $20$-prompt model (which includes the previous $10$ prompts plus \textit{``Make it look like a panda''}, \textit{``Make it look like made of bronze''}, \textit{``Turn it into blue''}, \textit{``Turn it into yellow''}, \textit{``Turn it into pink''}, \textit{``Turn it into green''}, \textit{``Turn it into red''}, \textit{``Add a snowboard to it''}, \textit{``Add sunglasses to it''}, \textit{``Add a crown to it''}).
As shown in \Cref{fig:scale}, the performance decreases slightly when moving from a single prompt to more prompts.
However, the difference between $10$ prompts and $20$ prompts is marginal.
This indicates the potential of our \method to scale to tens of prompts and even arbitrary prompts as inputs.

\section{Additional ablation studies}
\label{sec:ablation2}

\begin{figure}
\centering
\includegraphics[width=\linewidth]{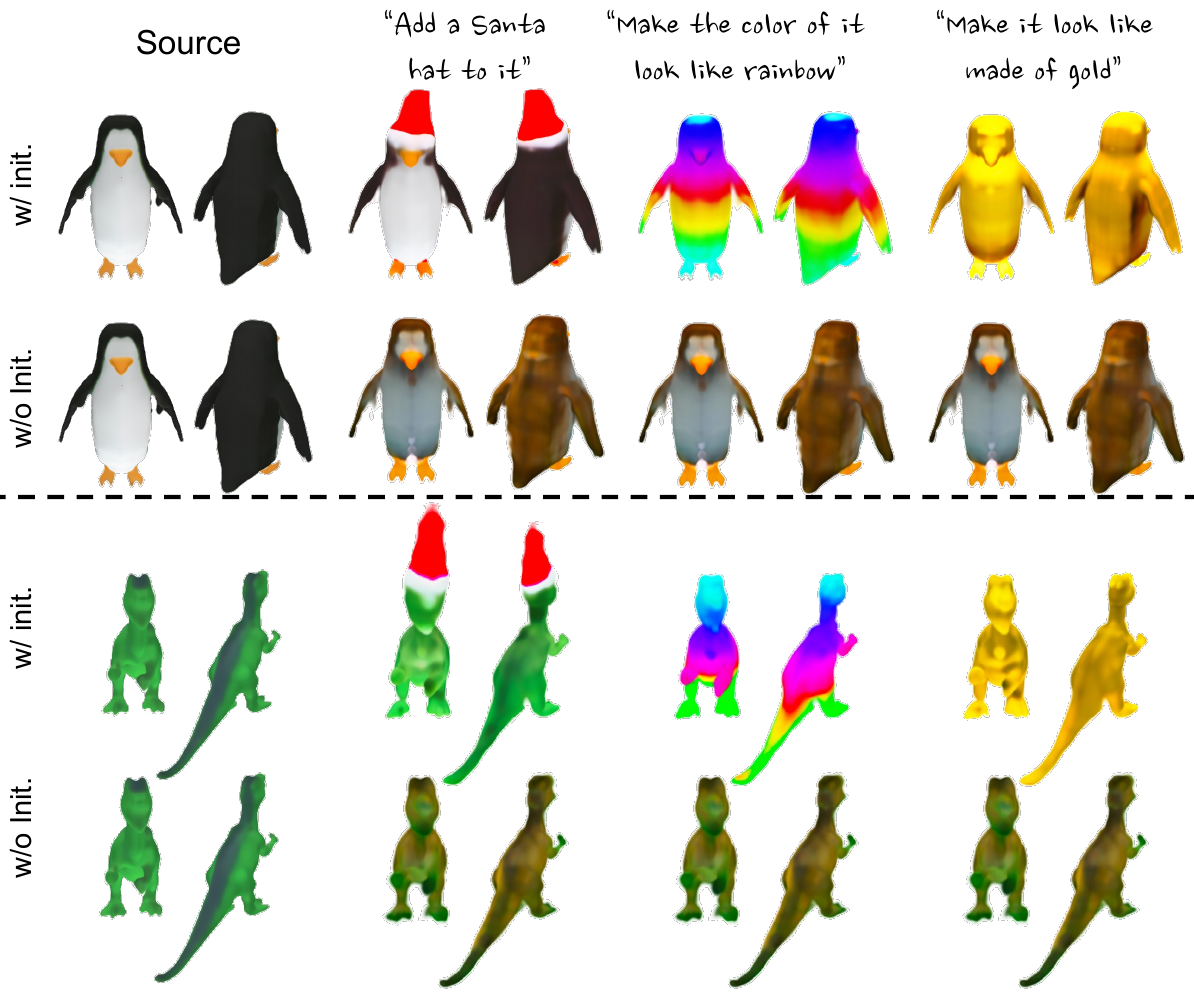}
\vspace{-0.6cm}
\caption{\textbf{Comparison with different initialisation method.} We use the multi-prompt setting in our experiments. ``w/ init.'' denotes initialisation using the pre-trained weights of the Shap-E text-to-3D diffusion model, and ``w/o init.'' indicates random initialisation. With random initialisation, the network loses the ability to distinguish across different prompts and produces similar results despite different instructions.}%
\label{fig:initialization}
\vspace{-0.cm}
\end{figure}

\paragraph{Network initialisation.} 
We compare the editing results of \method \textit{with} and \textit{without} the pre-trained weights of the text-to-3D Shap-E diffusion network quantitatively and qualitatively under the multi-prompt setting, \ie, Ours (Multi-prompt).
As shown in \Cref{fig:initialization}, the editor trained with Shap-E initialisation can successfully generate different effects given different prompts.
However, if instead we randomly initialise the network, the editor is unsuccessful, yielding similar outputs regardless of the prompt.
We assume this is because the model initialised with Shap-E pre-trained weights inherits its partial abilities to understand the natural language, and the randomly initialised one reaches a local optimum, ignoring the textual instructions.

\begin{figure}
\centering
\includegraphics[width=\linewidth]{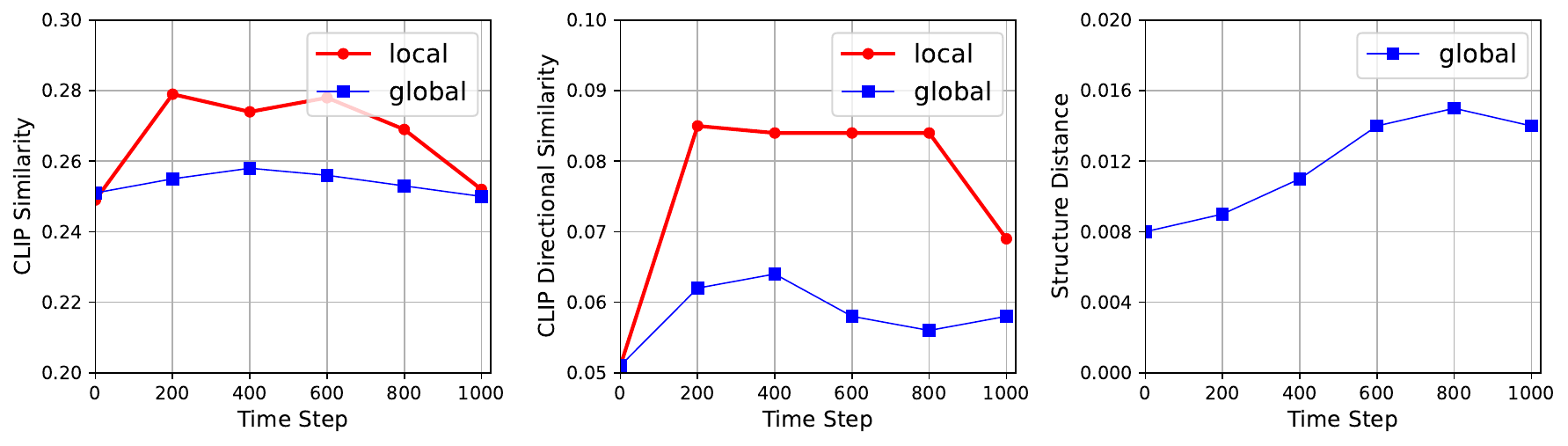}
\vspace{-0.55cm}
\caption{\textbf{Ablation study on the timestep $\tau$ for $\sigma_{\tau}$.} We analyse the level of noise introduced to the input, with a large timestep $\tau$ corresponding to a large noise. As the timestep $\tau$, thereby the noise level, increases, the Structure Distance rises consistently. When the input is the original latent ($\tau = 0$) or has very large noise ($\tau \to 1024$), we observe degraded performance in both CLIP similarity score and CLIP directional similarity score.}%
\label{fig:sigma}
\vspace{-0.cm}
\end{figure}

\paragraph{Effects of $\sigma_{\tau}$.}
Next, we study the value $\sigma_{\tau}$ that is used when noising the source latent to be able to initialise the editor with the pre-trained weights of the Shap-E diffusion model. 
To keep the full information and details of the original 3D asset, we follow \cite{brooks2023ip2p} and concatenate the noised latent with the original latent $(\sigma_\tau \br^s + \alpha_\tau \beps, \br^s)$.
Here, $\sigma_{\tau}$ is a hyperparameter that controls the information we keep from the original latent in the noised counterpart.
A smaller $\sigma_{\tau}$ means we keep less information.
The $\sigma_{\tau}$ value in the main text corresponds to $\tau = 200$ in the original total $1024$ Shap-E diffusion steps.
A higher $\tau$ corresponds to a smaller $\sigma_{\tau}$, and when $\tau = 1024$, the noised input can be considered as a random Gaussian noise.
As illustrated in the \Cref{fig:sigma}, time steps in the range $[200, 600]$ result in only marginal differences in performance.
A large noise or no noise will lead to a drop in the CLIP similarity and CLIP directional similarity scores.
Increasing the noise also leads to a larger Structure Distance.

\begin{figure}
\centering
\hspace{-0.2cm}
\includegraphics[width=\linewidth]{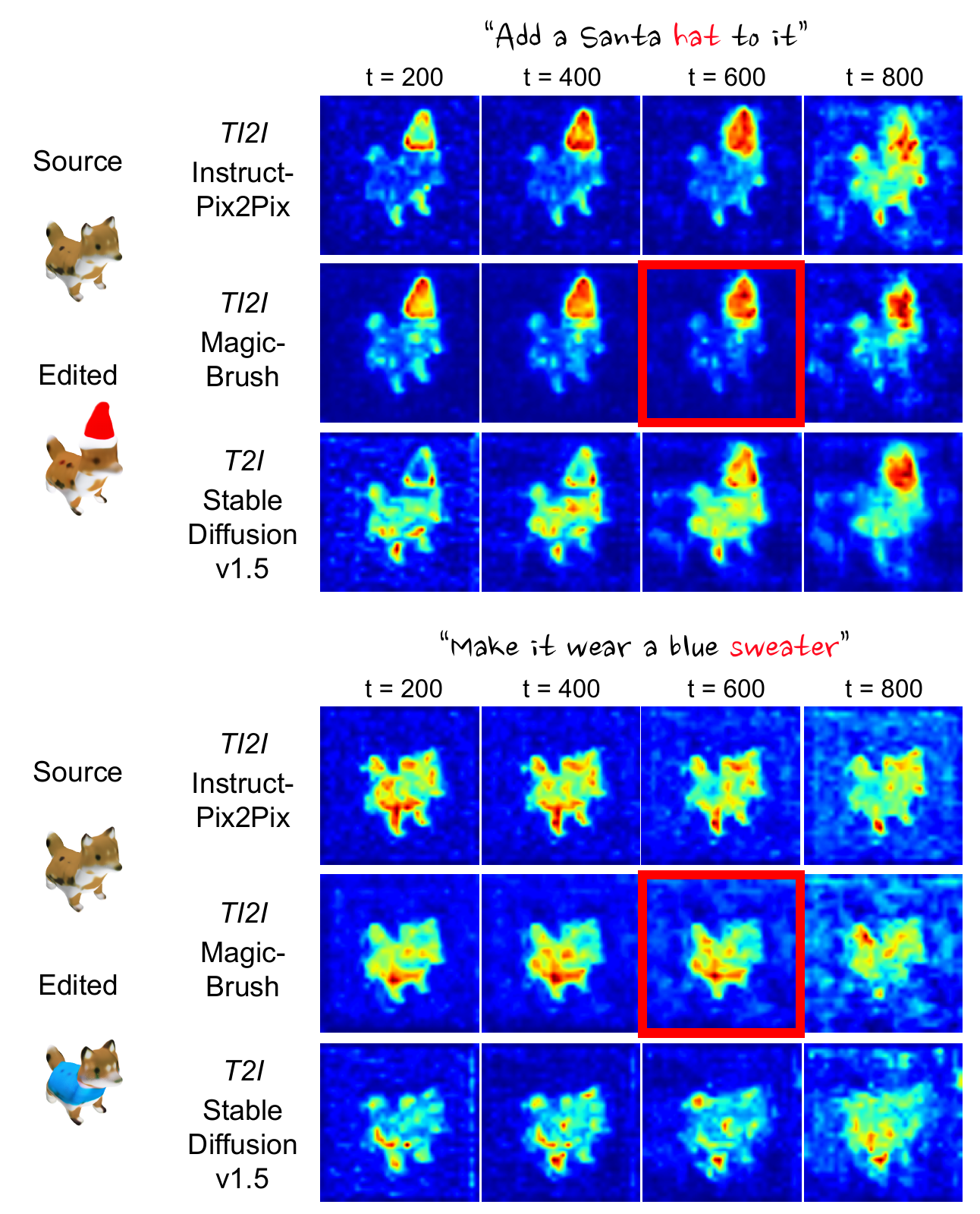}
\vspace{-0.1cm}
\caption{\textbf{Visualisation of cross-attention maps} that correspond to particular attention word tokens (labeled in \textcolor{red}{red}). These maps are extracted at different timesteps ($t \in [0, 1000] $) from various pre-trained 2D models (including InstructPix2Pix, MagicBrush, and Stable Diffusion v1.5). In this work, we consider the default cross-attention maps at $t=600$ from MagicBrush, which are indicated by the red boxes.}%
\label{fig:attn-map}
\vspace{-0.2cm}
\end{figure}

\paragraph{Choice of cross-attention maps.} 
To accurately estimate the local editing region, we assess the quality of cross-attention maps extracted at various timesteps from several pre-trained 2D models, including InstructPix2Pix~\cite{brooks2023ip2p}, MagicBrush~\cite{Zhang2023MagicBrush}, and Stable Diffusion v1.5~\cite{ho2020denoising}. As demonstrated in~\Cref{fig:attn-map}, most cross-attention maps either fail to disentangle the region of interest from the main object or suffer from excessive background noise. In contrast, the cross-attention map extracted from MagicBrush at $t = 600$ (indicated by red boxes) effectively highlights the region associated with the attention word tokens (\ie, \textit{``hat'', ``sweater''}). Therefore, we adopt this setting as the default in our experiments.

\section{Extended discussion on prior methods} 
\label{sec:ext_dicussion}
Our method differs from prior work employing test-time optimisation techniques in two main ways: (i) We use a direct, feed-forward approach to function on the 3D (latent) representation, unlike others that gradually optimise the 3D representation at test time to align it with a 2D prior. Our approach significantly reduces the inference time from tens of minutes to less than one second; (ii) Our method learns editing in a simpler, more structured latent space, avoiding the complexity of spaces like NeRF's weight matrix space. This simplification reduces learning difficulty and cost, allowing our model to generalise to novel objects at test time. 
Recently, EDNeRF~\cite{zheng2023editablenerf} tries to edit NeRFs that are trained on the latent space of Stable Diffusion~\cite{ho2020denoising}. The loss changes from the image space to the VAE latent space of Stable Diffusion. In this context, the use of the term ``latent'' is different from ours since it still requires NeRF as a representation of the 3D model and test-time optimisation.

Another key difference in our work is the use of complementary 2D diffusion priors for the training objectives. Other methods, such as IN2N~\cite{instructnerf2023}, Vox-E \cite{sella2023vox}, Instruct 3D-to-3D~\cite{kamata2023instruct} typically distill knowledge from \textit{one} network (\eg, Stable Diffusion~\cite{ho2020denoising} for Vox-E~\cite{brooks2023ip2p} and InstructPix2Pix~\cite{brooks2023ip2p} for IN2N and Instruct 3D-to-3D) with different regularisation due to different 3D representations.

As shown in our ablation studies in the main text, distilling from only one network usually inherits the drawbacks of the original 2D model, such as the inability to edit locally or preserve the original appearance and structure. Instead, one can distill from multiple 2D editors to overcome these pitfalls and achieve better editing results.

Finally, we also experiment with the training objective of IN2N, \ie, editing images directly and updating our editor function with a photometric loss.
However, this led to divergence, likely due to the greater inconsistency introduced by training with multiple instances compared to optimising a single NeRF.

\begin{figure}
\centering
\hspace{-0.2cm}
\includegraphics[width=\linewidth]{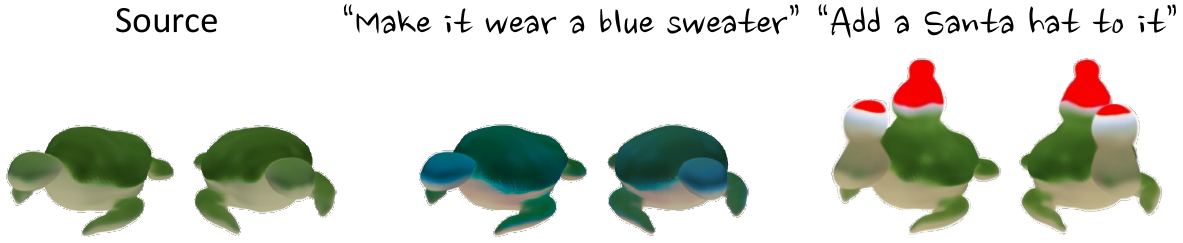}
\vspace{-0.1cm}
\caption{\textbf{Failure case.} When encountering a new class, such as \textit{``turtle''}, which significantly differs from those in the training dataset, our model struggles to identify the correct position for local editing.}
\label{fig:failure}
\vspace{-0.1cm}
\end{figure}
\section{Failure case}
\label{sec:failure_case}
In \Cref{fig:failure}, we present two failure cases. These occur particularly when the model encounters an unseen class that significantly differs from the classes the editor was trained on.
This disparity leads to difficulties in accurately determining the position for local editing, ultimately resulting in editing failures.
We conjecture that such failure cases can be eliminated by training the editor on an even larger number of object categories.